\documentclass[letterpaper]{article} 
\usepackage{aaai24}  
\usepackage{times}  
\usepackage{helvet}  
\usepackage{courier}  
\usepackage[hyphens]{url}  
\usepackage{graphicx} 
\usepackage{natbib}  
\usepackage{caption} 
\usepackage{algorithm}
\usepackage{algorithmic}

\usepackage{newfloat}
\usepackage{listings}
\DeclareCaptionStyle{ruled}{labelfont=normalfont,labelsep=colon,strut=off} 
\floatstyle{ruled}
\newfloat{listing}{tb}{lst}{}
\floatname{listing}{Listing}

\usepackage{graphicx}
\usepackage{latexsym}

\usepackage{times}
\usepackage{soul}
\usepackage{url}
\usepackage{graphicx}
\usepackage{amsmath}
\usepackage{amssymb}
\usepackage{amsthm}
\usepackage{booktabs}
\usepackage{subfigure}
\usepackage{diagbox}
\usepackage[switch]{lineno}
\usepackage{etoolbox}

\newtheorem{proposition}{Proposition}
\title{Learning Diverse Risk Preferences in Population-based Self-play}
\author {
    Yuhua Jiang\equalcontrib,
    Qihan Liu\equalcontrib,
    Xiaoteng Ma,
    Chenghao Li, 
    Yiqin Yang, \\
    Jun Yang\footnotemark[2],
    Bin Liang,
    Qianchuan Zhao\thanks{Corresponding authors.}}
\affiliations {
    Department of Automation, Tsinghua University \\
    \{jiangyh22, lqh20, ma-xt17, lch18, yangyiqi19\}@mails.tsinghua.edu.cn \\
    \{yangjun603, bliang, zhaoqc\}@tsinghua.edu.cn
    }

\begin{document}

\maketitle

\begin{abstract}
Among the remarkable successes of Reinforcement Learning (RL), self-play algorithms have played a crucial role in solving competitive games. However, current self-play RL methods commonly  optimize the agent to maximize the expected win-rates against its current or historical copies, resulting in a limited strategy style and a tendency to get stuck in local optima. 
To address this limitation, it is important to improve the diversity of policies, allowing the agent to break stalemates and enhance its robustness when facing with different opponents.  In this paper, we present a novel perspective to promote diversity by considering that agents could have diverse risk preferences in the face of uncertainty.
To achieve this, we introduce a novel reinforcement learning algorithm called Risk-sensitive Proximal Policy Optimization (RPPO), which smoothly interpolates between worst-case and best-case policy learning, enabling policy learning with desired risk preferences. Furthermore, by seamlessly integrating RPPO with population-based self-play, agents in the population optimize dynamic risk-sensitive objectives using experiences gained from playing against diverse opponents. 
Our empirical results demonstrate that our method achieves comparable or superior performance in competitive games and, importantly, leads to the emergence of diverse behavioral modes. Code is available at \url{https://github.com/Jackory/RPBT}.

\end{abstract}

\section{Introduction}
Reinforcement Learning (RL) has witnessed significant advancements in solving challenging decision problems, particularly in competitive games such as Go~\cite{silver2016mastering}, Dota~\cite{openai2018openai}, and StarCraft~\cite{vinyals2019grandmaster}. 
One of the key factors contributing to these successes is self-play, where an agent improves its policy by playing against itself or its previous policies. However, building expert-level AI solely through model-free self-play remains challenging, as it requires no access to a dynamic model or the prior knowledge of a human expert. One of the difficulties is that agents trained in self-play only compete against themselves, leading to policies that often become stuck in local optima and struggle to generalize to different opponents~\cite{bansal2018emergent}. 

In this study, we focus on the problem of training agents that are both robust and high-performing against any type of opponent. We argue that the key to addressing this problem lies in producing a diverse set of training opponents, or in other words, learning diverse strategies in self-play. A promising paradigm to address this problem is population-based approaches~\cite{vinyals2019grandmaster,jaderberg2019humanlevel}, where a collection of agents is gathered and trained against each other. However, the diversity achieved in such methods mainly stems from various hyperparameter setups and falls short. Recent research has increasingly focused on enhancing population diversity~\cite{lupu2021trajectory,zhao2022maximum,parker-holder2020effective,wu2023qualitysimilar,liu2021towards}. These methods introduce diversity as a learning objective and incorporate additional auxiliary losses on a population-wide scale. Nonetheless, this population-wide objective increases implementation complexity and necessitates careful hyperparameter tuning to strike a balance between diversity and performance.

Here, we propose a novel perspective on introducing diversity within the population: agents should possess diverse risk preferences. In this context, risk refers to the uncertainty arising from stochastic transitions within the environment, while risk preference encompasses the agent's sensitivity to such uncertainty, typically including risk-seeking, risk-neutral, and risk-averse tendencies. Taking inspiration from the fact that humans are risk-sensitive~\cite{tversky1992advances}, we argue that the lack of diversity in self-play arises from agents solely optimizing their expected winning rate against opponents without considering other optimization goals that go beyond the expectation, such as higher-order moments of the winning rate distribution. Intuitively, the population of agents should be diverse, including conservative (risk-averse) agents that aim to enhance their worst performance against others, as well as aggressive (risk-seeking) agents that focus on improving their best winning record.

To achieve risk-sensitive learning with minimal modifications, we introduce the \emph{expectile Bellman operator}, which smoothly interpolates between the worst-case and best-case Bellman operators. Building upon this operator, we propose \emph{Risk-sensitive Proximal Policy Optimization} (RPPO), which utilizes a risk level hyperparameter to control the risk preference during policy learning. To strike a balance between bias and variance in policy learning, we extend the expectile Bellman operator to its multi-step form and adapt the Generalized Advantage Estimation (GAE)~\cite{schulman2018highdimensional}. It is important to highlight that this extension is non-trivial due to the nonlinearity of the expectile Bellman operator.

RPPO offers a simple and practical approach that can be easily integrated into existing population-based self-play frameworks. We provide an implementation called RPBT, where a population of agents is trained using RPPO with diverse risk preference settings, without introducing additional population-level optimization objectives. The risk preferences of the agents are automatically tuned using exploitation and exploration techniques drawn from Population-based Training (PBT)~\cite{jaderberg2017population}. By gathering the experience of agents competing against each other in the diverse population, RPBT effectively 
 addresses the issue of overfitting to specific types of opponents.

    

    
    

\section{Related Work}
\paragraph{Self-play RL}
Training agents in multi-agent games requires instantiations of opponents the in environment. A solution could be self-play RL, where an agent is trained by playing against its own policy or its past versions. Self-play variants have proven effective in some multi-agent games~\cite{tesauro1995temporal,silver2016mastering,openai2018openai,vinyals2019grandmaster,jaderberg2019humanlevel}. A key advantage of self-play is that the competitive multi-agent environment provides the agents with an appropriate curriculum~\cite{bansal2018emergent,liu2019emergent,baker2020emergenta}, which facilitates the emergence of complex and interesting behaviors. However, real-world games often involve non-transitivity and strategic cycles\cite{czarnecki2020real}, and thus self-play cannot produce agents that generalize well to different types of opponents. Population-based self-play~\cite{jaderberg2019humanlevel,garnelo2021pick,
yu2023learning,strouse2021collaborating} improves self-play by training a population of agents, all of whom compete against each other. While population-based self-play is able to gather a substantial amount of match experience to alleviate the problem of overfitting to specific opponents, it still requires techniques to introduce diversity among agents to stabilize training and facilitate robustness~\cite{jaderberg2019humanlevel}. 

\paragraph{Population-based Diversity}
Population-based methods demonstrate a strong connection with evolutionary algorithms~\cite{mouret2015illuminating}. These population-based approaches have effectively addressed black-box optimization challenges~\cite{loshchilov2016cmaes}. Their primary advantages encompass the ability to obtain high-performing hyperparameter schedules in a single training run, which leads to great performance across various environments~\cite{liu2019emergent,li2019generalized,espeholt2018impala}.
Recently, diversity across the population has drawn great interests~\cite{shen2020generating, khadka2019collaborative,liu2021towards,liu2022unified,wu2023qualitysimilar}. Reward randomizations~\cite{tang2021discovering,yu2023learning} were employed to discover diverse policies in multi-agent games. Other representative works formulated individual rewards and diversity among agents into a multi-objective optimization problem. More specifically, DvD ~\cite{parker-holder2020effective} utilizes the determinant of the kernel matrix of action embedding as a population diversity metric. TrajeDi~\cite{lupu2021trajectory} introduces trajectory diversity by approximating Jensen-Shannon divergence with action discounting kernel. MEP~\cite{zhao2022maximum} maximizes a lower bound of the average KL divergence within the population. Instead of introducing additional diversity-driven objectives across the entire population, our RPBT approach trains a population of agents with different risk preferences. Each individual agent maximizes its own returns based on its specific risk level, making our method both easy to implement and adaptable for integration into a population-based self-play framework.

\paragraph{Game theory}  Fictitious play~\cite{brown1951fp} and double oracle\cite{mcmahan2003doubleoracle} have been studied to achieve approximate Nash equilibrium in normal-form games. 
Fictitious self-play (FSP)\cite{heinrich2015fictitious} extends fictitious play to extensive-form games. Policy-space response oracle (PSRO)~\cite{lanctot2017unified} is a natural generalization of double oracle, where the choices become the policies in meta-games rather than the actions in games. 
PSRO is a general framework for solving games, maintaining a policy pool and continuously adds the best responses. Our methods fall under PSRO framework at this level. However, in practice, PSRO necessitates the computation of the meta-payoff matrix between policies in the policy pool, which is computationally intensive~\cite{mcaleer2020pipeline} in real-world games since the policy pool is pretty large. Various improvements~\cite{balduzzi2019open,perez2021modelling,liu2021towards} have been made upon PSRO by using different metrics based on the meta-payoffs to promote diversity. However, most of these works have confined their experiments to normal-form games or meta-games.

\paragraph{Risk-sensitive RL} 
Our risk-sensitive methods draw from risk-sensitive and distributional RL, with comprehensive surveys~\cite{bellemare2023distributional} available. Key distributional RL studies~\cite{bellemare2017distributional} highlight the value of learning return distributions over expected returns, enabling approximation of value functions under various risk measures like Wang~\cite{muller1997integral} and CVaR~\cite{rockafellar2000optimization,chow2014algorithms,qiu2021rmix} for generating risk-averse or risk-seeking policies. However, these methods' reliance on discrete samples for risk and gradient estimation increases computational complexity.
Sampling-free methods~\cite{tang2019worst,yang2021wcsac,ying2022towards} have been explored for CVaR computation, but CVaR focuses solely on risk aversion, neglecting best-case scenarios, which may not align with competition objectives. A risk-sensitive RL algorithm~\cite{deletang2021model} balances risk aversion and seeking, but assumes gaussian data generation. In contrast, our RPPO algorithm requires no data assumptions and minimal code modifications on PPO to learn diverse risk-sensitive policies.


\section{Preliminary}
\paragraph{Problem Definition}
We consider fully competitive games, which can be regarded as Markov games~\cite{littman1994markov}. A Markov game for $N$ agents is a partially observable Markov decision process (POMDP), which can be defined by: a set of states $\mathcal{S}$, a set of observations $\mathcal{O}^1, \cdots, \mathcal{O}^N$ of each agent, a set of actions of each agent $\mathcal{A}^1, \cdots, \mathcal{A}^N$, a transition function  $p(s^{\prime} |s,a_1,\cdots,a_n)$: $\mathcal{S} \times \mathcal{A}^1 \times \cdots \mathcal{A}^N \rightarrow \Delta({\mathcal{S}})$ determining distribution over next states, and a reward function for each agent $r^i: \mathcal{S} \times \mathcal{A}^i \times \mathcal{A}^{-i} \times \mathcal{S} \rightarrow \mathbb{R}$. Each agent chooses its actions based on a stochastic policy $\pi_{\theta_i}: \mathcal{O}^i \rightarrow \Delta({\mathcal{A}})$, where $\theta_i$ is the policy parameter of agent $i$.

In the self-play setting, we have control over a single agent known as the main agent, while the remaining agents act as opponents. These opponents are selected from the main agent's checkpoints. The main agent's primary objective is to maximize its expected returns, also referred to as discounted cumulative rewards, denoted as $\mathbb{E}[\sum_{t=0}^T \gamma^t r_t^i$], where $\gamma$ represents the discount factor, and $T$ signifies the time horizon. By considering the opponents as part of the environmental dynamics, we can view the multi-agent environment as a single-agent stochastic environment from the main agent's perspective. Consequently, we can employ single-agent RL methods such as PPO to approximate the best response against all opponents. However, it should be noted that the opponents are sampled from the main agent, which itself lacks diversity as it is generated by single-agent RL methods. In this study, we aim to construct a strong and robust policy, and generating diverse policies serves this goal.

\section{Risk-sensitive PPO} \label{sec:rppo}
In this section, we present our novel RL algorithm, Risk-sensitive Proximal Policy Optimization (RPPO), which involves minimal modifications to PPO. The algorithm is shown in Algorithm \ref{algo:rppo}. RPPO utilizes an expectile Bellman operator that interpolates between a worst-case Bellman operator and a best-case Bellman operator. This allows learning to occur with a specific risk preference, ranging from risk-averse to risk-seeking. Additionally, we extend the expectile Bellman operator into a multi-step form to balance bias and variance in RPPO. The theoretical analysis of the expectile Bellman operator and its multi-step form, as well as a toy example demonstrating the potential of RPPO to learn risk preferences, are presented in the following subsections.

\subsection{Expectile Bellman operator}
\label{sec:operator} 
 Given a policy $\pi$ and a risk level hyperparamemter $\tau \in (0,1)$, we consider expectile Bellman operator defined as follows:
\begin{equation}
\begin{aligned}
    \mathcal{T}_{\tau}^{\pi} V(s) := V(s) + 2\alpha \mathbb{E}_{a} \mathbb{E}_{s^\prime} \left[\tau[\delta]_{+} + (1-\tau)[\delta]_{-}\right],
\end{aligned}
\end{equation}
where $\alpha$ is the step size which we set to $\frac{1}{2 \max \{\tau, 1-\tau\}}$, $\delta$ refers to $\delta(s,a,s^\prime) = r(s,a,s^\prime) + \gamma V(s^{\prime}) - V(s)$ which is one-step TD error, $[\cdot]_+ = \max(\cdot,0)$ and $[\cdot]_{-}=\min(\cdot,0)$. This operator draws inspiration from expectile statistics~\cite{newey1987asymmetric,rowland2019statisticsb,ma2022offline}, and thus the name.
We can see that the standard Bellman operator is a special case of expectile Bellman operator when $\tau=1/2$. 
We have the following theoretical properties to guide the application of expectile Bellman operator in practice. Please refer to the Appendix \ref{app:theory} for the proof.


\begin{proposition}
\label{proposition:one_contraction}
For any $\tau \in(0,1), \mathcal{T}_\tau^\pi$ is a $\gamma_\tau$-contraction, where $\gamma_\tau=1-2 \alpha(1-\gamma) \min \{\tau, 1-\tau\}$.
\end{proposition}

Proposition~\ref{proposition:one_contraction} guarantees the convergence of the value function in the policy evaluation phase.

\begin{proposition}
\label{proposition:one_monotoic}
Let $V_{\tau}^*$ denotes the fixed point of $\mathcal{T}_\tau^\pi$. For any $\tau, \tau^{\prime} \in(0,1)$, if $\tau^{\prime} \geq \tau$, we have $V_{\tau^{\prime}}^*(s) \geq V_\tau^*(s), \forall s \in S$.
\end{proposition}

Proposition~\ref{proposition:one_monotoic} guarantees the fixed point of expectile Bellman operator is monotonic with respect to $\tau$.

\begin{proposition}
\label{proposition:one_risk}
Let $V^*_{\tau}$, $V^*_{best}$, and $V^{*}_{worst}$ respectively denote the fixed point of expectile Bellman operator, best-case Bellman operator and worst-case Bellman operator. We have 
\begin{equation}
    V_{\tau}^*= 
    \begin{cases}
        V_{worst}^* & \text { if } \tau \rightarrow 0 \\ 
        V_{best}^* & \text { if } \tau \rightarrow 1.
    \end{cases}
\end{equation}
\end{proposition}

The worst-case Bellman operator and best-case Bellman operator are defined by:
\begin{equation}
\begin{aligned}
    \mathcal{T}_{best}V(s):=\max_{a,s^\prime}[R(s,a,s^\prime)+\gamma V(s^{\prime})], \\
    \mathcal{T}_{worst}V(s)=\min_{a,s^\prime}[R(s,a,s^\prime)+\gamma V(s^{\prime})].
\end{aligned}
\end{equation}
It is important to highlight the difference between best-case Bellman operator and  Bellman optimal operator ($\mathcal{T}V(s):=\max_{a}[\sum_{s^{\prime}}(R(s,a,s^{\prime})+\gamma V(s^{\prime})]$). Best-case Bellman operator takes stochastic transitions into account, which is the main source of risk in competitive games when confronting unknown actions of different opponents.  

Proposition~\ref{proposition:one_risk} guarantees expectile Bellman operator approaches best-case Bellman operator as risk level $\tau$ approaches 1, and approaches worst-case Bellman operator as $\tau$ approaches 0. 

Combing Propostion~\ref{proposition:one_monotoic} and Proposition~\ref{proposition:one_risk}, we observe that expectile Bellman operator can be used to design an optimization algorithm whose objective is the interpolation between the best-case and the worst-case objective. When $\tau=1/2$, the objective is equivalent to expected returns, which is the scenario of risk-neutral. And risk level $\tau < 1/2$ and $\tau > 1/2$ represent the risk-averse and risk-seeking cases, respectively. As $\tau$ varies, the learning objective varies, and hence diverse risk perferences arise.

We can see the potential of using the expectile Bellman operator for designing risk-sensitive RL algorithms. PPO, widely utilized in self-play, is favored for its ease of use, stable performance, and parallel scalability. By extending PPO with the expectile Bellman operator, we introduce the RPPO algorithm. Moreover, generalizing this operator to other RL algorithms is not difficult. In practice, we define the advantage function as
\begin{equation}
    A^\pi_\tau(s, a) := 2\alpha\mathbb{E}_{ s^{\prime}}\left[
    \tau\left[\delta\right]_+ + (1-\tau)[\delta]_-\right],
\end{equation}

With a batch of data $\{(s_t, a_t, r_t, s_{t+1})\}_{t=0}^{T-1}$ collected by $\pi$, we train the policy with the clip variant PPO loss,
\begin{equation}
\begin{aligned}\label{equ:policy_loss}
    \mathcal{L}_\pi^{\rm CLIP}(\theta) =
    \frac{1}{T}\sum_{t=0}^{T-1} &\left[
    \min \left(\omega_t(\theta) \hat{A}_\tau(s_t, a_t), \right.\right.
    \\
    &\phantom{=\quad}\left.\left.\text{clip}( \omega_t(\theta),
    1-\varepsilon,
    1+\varepsilon)\hat{A}_\tau(s_t, a_t) \right) \right],
\end{aligned}
\end{equation}
where $\omega_t(\theta) = \frac{\pi_\theta(a_t \mid s_t)}{\pi(a_t \mid s_t)}$ is the importance sampling ratio. Meanwhile, we train the value network with mean squared loss,
\begin{equation} \label{equ:value_loss}
    \mathcal{L}_V(\phi) = \frac{1}{2T} \sum_{t=0}^{T-1} (V_\phi(s_t) - \hat{V}_{\tau, t})^2,
\end{equation}
where $\phi$ is the parameter of the value network, and $\hat{V}_{\tau,t}:= V_\phi(s_t)+ \hat{A}_\tau(s_t, a_t)$ is the target value.

\begin{algorithm}[tb] 
    \caption{Risk-sensitive PPO (RPPO)}
    \label{algo:rppo}
    \textbf{Input}: initial network parameter $\theta$, $\phi$, horizon $T$,  update epochs $K$, risk level $\tau$.
    \begin{algorithmic}[1]
        \FOR {$i=1,2,\cdots$} 
        \STATE Collect trajectories by running policy $\pi_{\theta_{old}}$ in the environment for $T$ time steps.
        \STATE Compute $\lambda$-variant returns $\hat{V}_{\tau,\lambda}$ according to Eq.\ref{equ:lambda_return}
        \STATE Compute advantages $\hat{A}_{\tau, \lambda}$ according to Eq.\ref{equ:lambda_advantage}
        \FOR {$k=1,2,\cdots,K$}
        \STATE Update the $\theta$ by maximizing the surrogate function with Eq.\ref{equ:policy_loss}
        via some gradient algorithms.
        \STATE Update the $\phi$ by minimizing mean-squared error with Eq.\ref{equ:value_loss} via some gradient algorithms.
        \ENDFOR
        \STATE $\theta_{old}=\theta$
        \ENDFOR
    \end{algorithmic}
\end{algorithm}

\begin{figure*}
    \centering
    \subfigure[]{
    \includegraphics[height=0.15\textwidth]{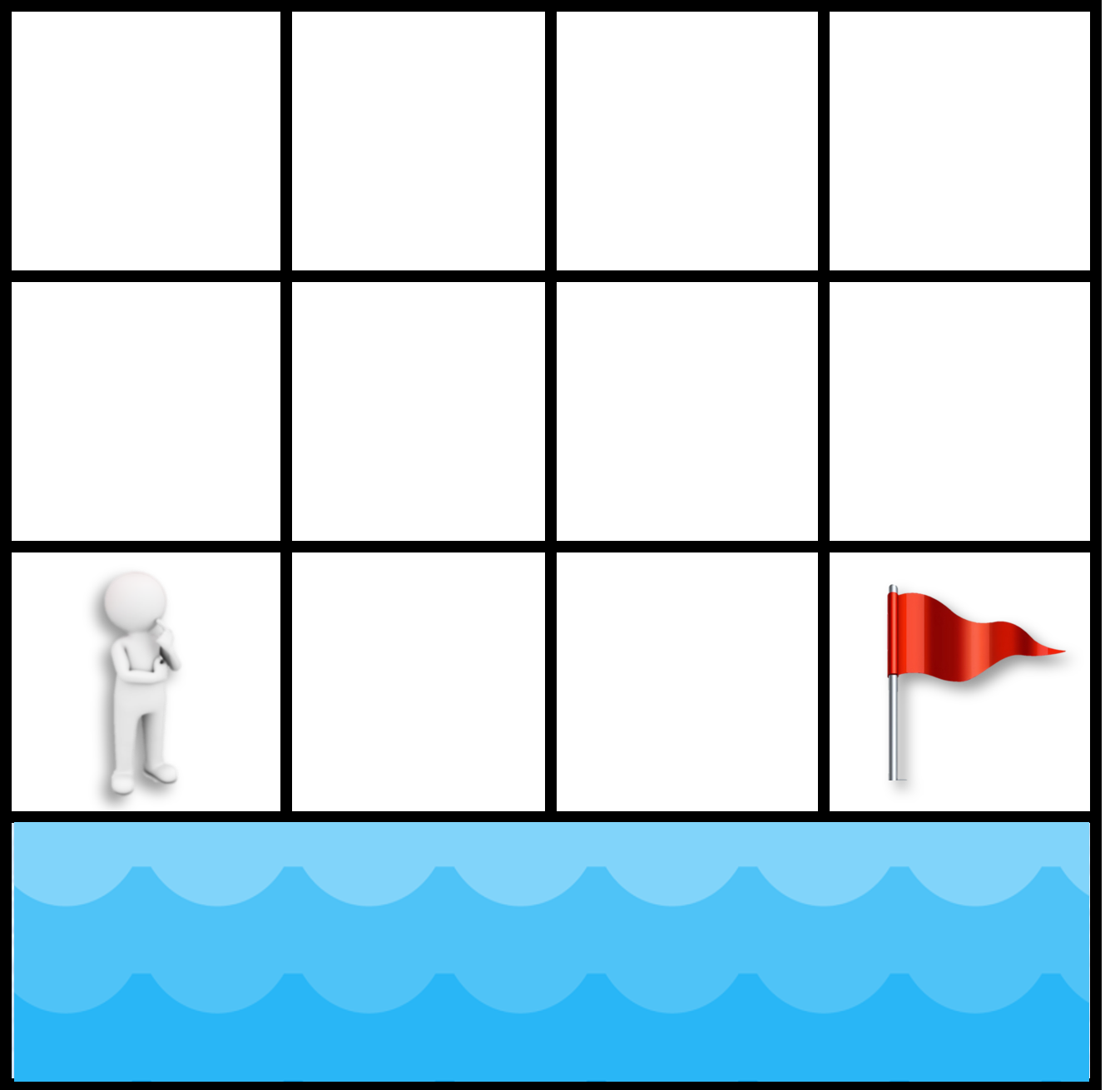}
    \label{fig:toy.a}
    }
    \subfigure[$\tau=0.2$]{
    \includegraphics[width=0.18\textwidth]{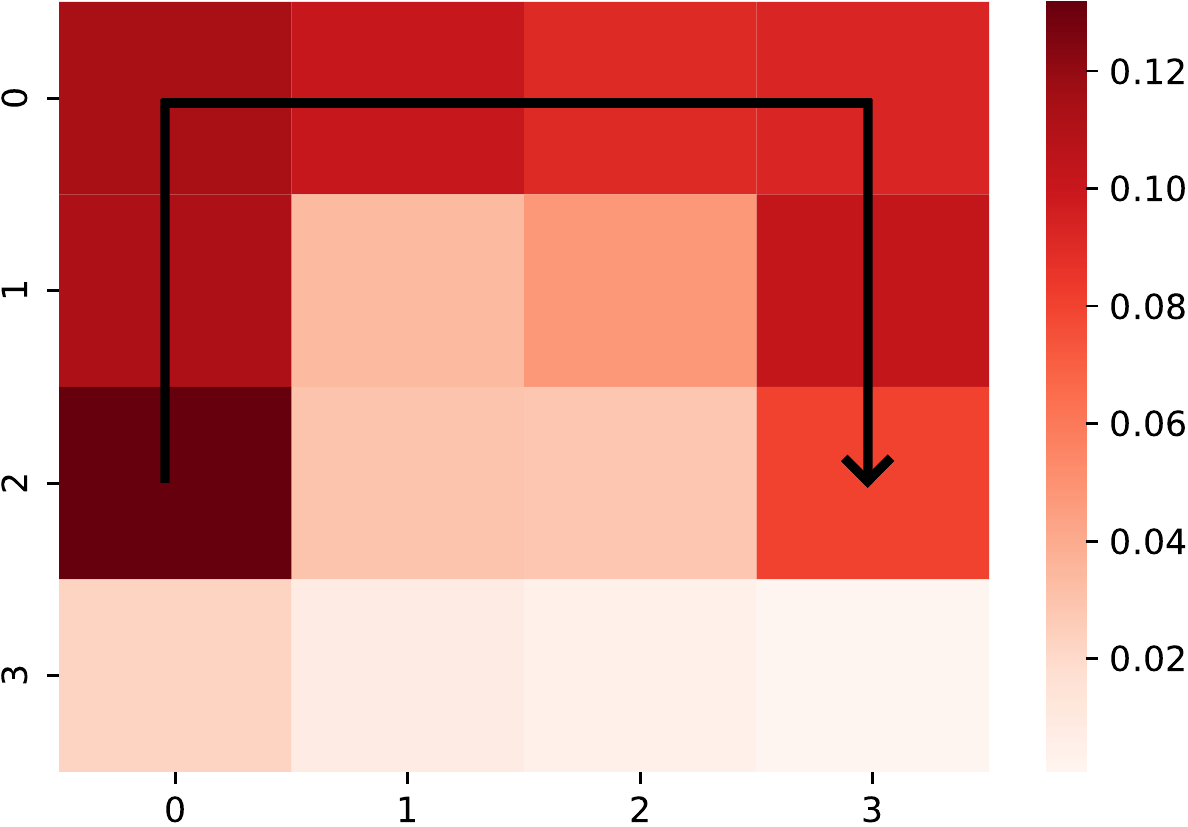}
    \label{fig:toy.risk_averse}
    }
    \subfigure[$\tau=0.5$]{
    \includegraphics[width=0.18\textwidth]{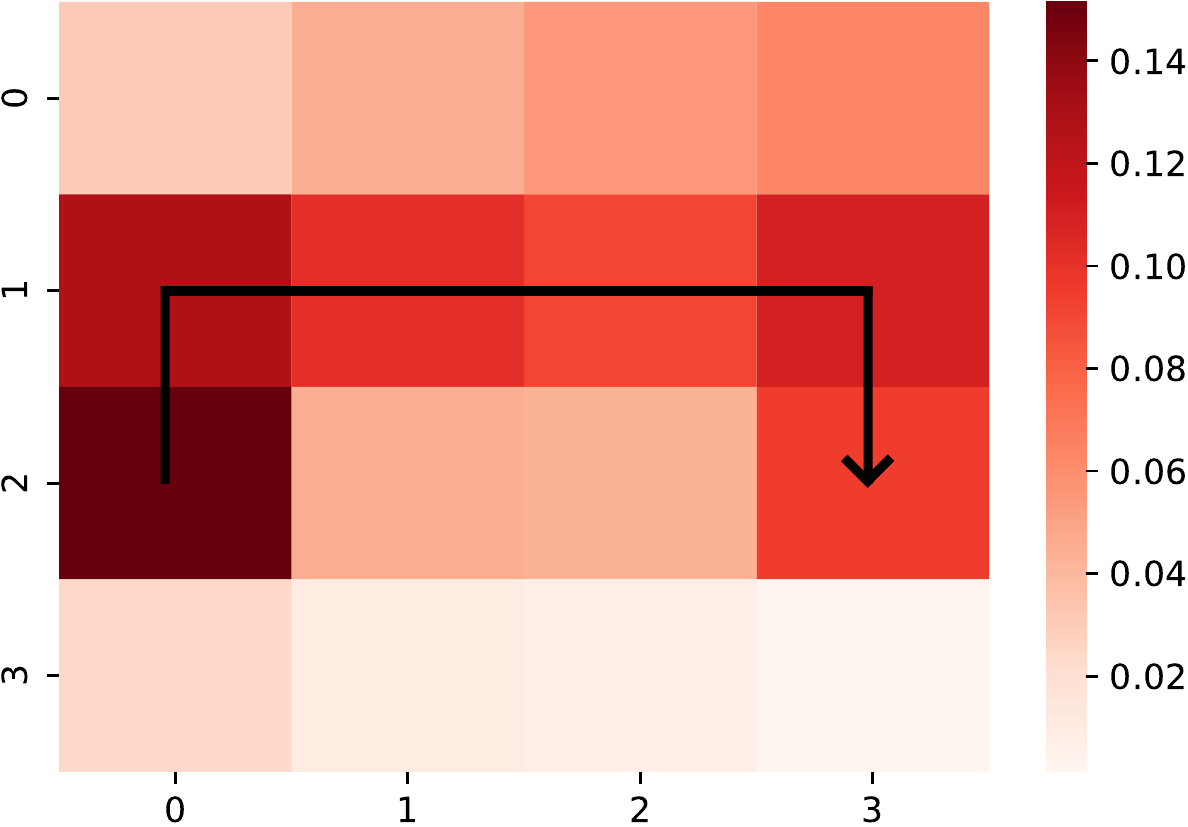}
    \label{fig:toy.risk_neutral}
    }
    \subfigure[$\tau=0.8$]{
    \includegraphics[width=0.18\textwidth]{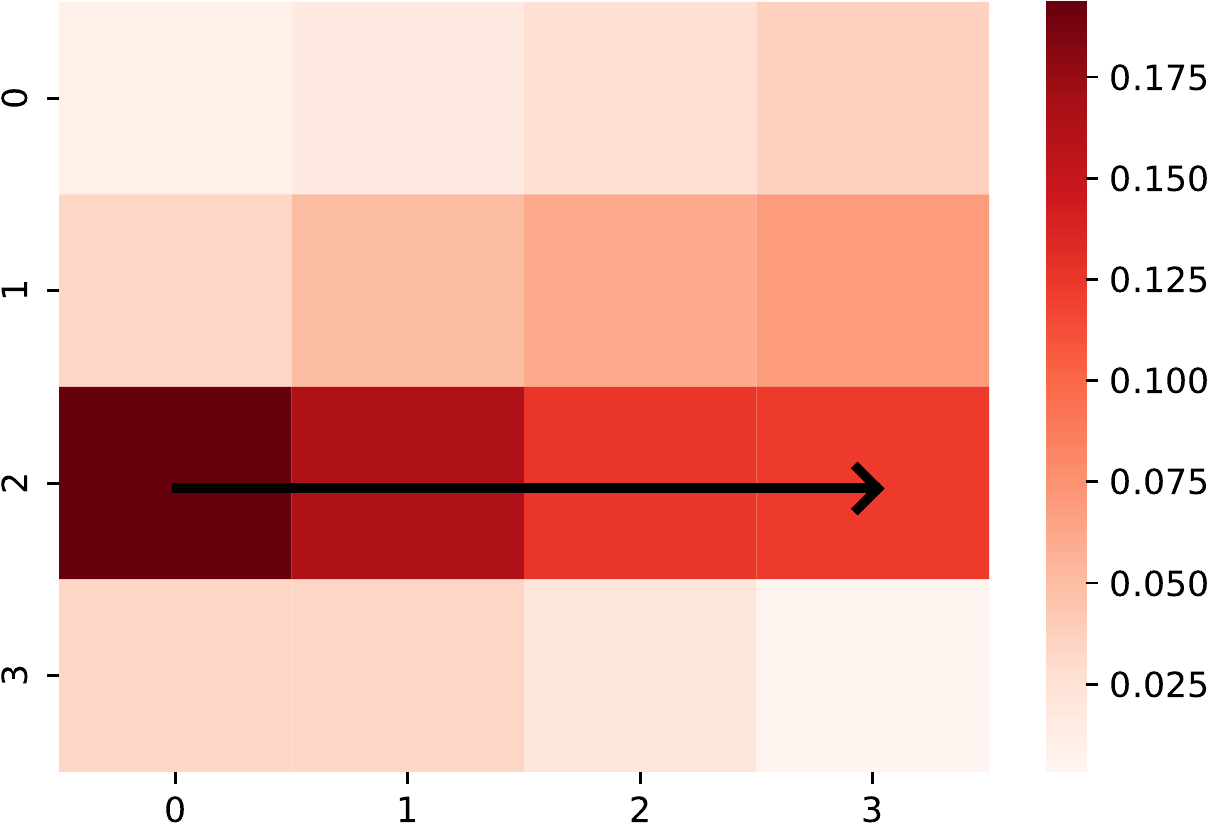}
    \label{fig:toy.risk_seeking}
    }
    \caption{Experiments of the toy example. (a) Illustration of the toy example. The task is to pick up the bonus located at the flag while avoiding the penalty of stepping into water in a $4\times4$ grid world. A strong wind pushes the agent into a random direction 50\% of the time. (b), (c) and (d) States visitation frequencies for the three policies that have different risk preferences. The black arrow indicates the deterministic policy when there is no wind.}
    \label{fig:toy}
\end{figure*}

\subsection{Multi-step expectile Bellman operator}
While RPPO is sufficient for providing decent risk preferences, we still require some techniques to balance the bias and variance when estimating the advantage function. The original PPO uses the Generalized Advantage Estimation (GAE) technique \cite{schulman2018highdimensional} to reduce variance by using an exponentially-weighted estimator of the advantage function, at the cost of introducing some bias. In this section, we extend GAE to RPPO along this line of work. However, it is non-trivial for direct incorporation of GAE into RPPO due to the \emph{non-linearity} of the expectile Bellman operator. To address this issue, we define the multi-step expectile Bellman operator as:

\begin{equation}
\mathcal{T}_{\tau, \lambda}^\pi V(s) := (1 - \lambda) \sum_{n=1}^{\infty} \lambda^{n-1} (\mathcal{T}_\tau^\pi)^n V(s).
\end{equation}

We can derive that multi-step expectile Bellman operator still remains the contraction and risk-sensitivity properties. Please refer to Appendix \ref{app:theory} for the proof.

\begin{proposition}\label{proposition:multi_contraction}
For any $\tau \in (0,1)$, $\mathcal{T}_{\tau, \lambda}^\pi$ is a $\gamma_{\tau,\lambda}$-contraction, where $\gamma_{\tau,\lambda} = \frac{(1 - \lambda)\gamma_\tau}{1 - \lambda \gamma_\tau}$.
\end{proposition}

\begin{proposition} \label{proposition:multi_monotoic}
Let $V_{\tau,\lambda}^*$ denote the fixed point of $\mathcal{T}_{\tau,\lambda}^\pi$. For any $\tau, \tau^{\prime} \in(0,1)$, if $\tau^{\prime} \geq \tau$, we have $V_{\tau^{\prime},\lambda}^*(s) \geq V_{\tau,\lambda}^*(s), \forall s \in S$.
\end{proposition}

\begin{proposition} \label{proposition:multi_risk}
Let $V^*_{\tau,\lambda}$ denote the fixed point of $\mathcal{T}^{\pi}_{\tau,\lambda}$, we have
\begin{equation}
    \lim _{\tau} V_{\tau,\lambda}^*= 
    \begin{cases}
        V_{worst}^* & \text { if } \tau \rightarrow 0 \\ 
        V_{best}^* & \text { if } \tau \rightarrow 1
    \end{cases}.
\end{equation}
\end{proposition}

Despite the fact that $\mathcal{T}_{\tau, \lambda}^\pi$ has the property of contraction, it is hard to be estimated with trajectories. Hence, we introduce another sample form operator
\begin{equation}
    \hat{\mathcal{T}}_\tau^{\pi, \hat V} V(s) := V(s) + 2\alpha\left[\tau [\hat \delta]_+ + (1-\tau)[\hat \delta ]_-\right],
\end{equation}
where $\hat \delta :=r(s,a,s^{\prime})+ \gamma \hat V(s') - V(s)$. Here $\hat V$ is an estimate of target value. When we choose $\hat V = V$ and take the expectation, we recover the $\mathcal{T}_\tau^{\pi}(s) = \mathbb{E}_{a\sim \pi(\cdot \mid s), s^{\prime} \sim p(\cdot \mid s, a)} \hat{\mathcal{T}}_\tau^{\pi, V} V(s)$. Furthermore, The multi-step sample form operator is
\begin{equation}\label{equ:lambda_return}
    \hat{V}^{\pi}_{\tau, \lambda}(s) = 
    (1 - \lambda) \sum_{n=1}^{\infty} \lambda^{n-1} \hat{\mathcal{T}}_\tau^{\pi, \hat V_n} V(s).
\end{equation}
where $\hat{V}_n(s) = \hat{\mathcal{T}}_\tau^{\pi, \hat{V}_{n-1}} V(s)$ and $\hat{V}_0(s) = V(s)$.
Finally, we estimate the advantage as 
\begin{equation}\label{equ:lambda_advantage}
    \hat{A}^\pi_{\tau, \lambda}(s, a) = 
    \hat{V}^\pi_{\tau, \lambda}(s) - V(s).
\end{equation}
When $\tau=1/2$, we recover the original form of GAE. However, if $\tau \neq 1/2$, we introduce extra bias\footnote{The fixed points of $\mathcal{T}_{\tau, \lambda}^\pi$ and $\mathcal{T}_{\tau}^\pi$ are the same one. However, the sample form operator introduces extra bias for value estimation, which means the fixed points of $\hat{\mathcal{T}}^\pi_{\tau, \lambda}$ and $\mathcal{T}^\pi_{\tau, \lambda}$ are not the same one. It is caused by the non-linearity of the operator with respect to the noise.}. The details of computing multi-step expectile Bellman operator in practice are provided in Appendix \ref{appendix:implement}.

With sample form multi-step expectile Bellman operator in place, we propose the whole RPPO algorithm, as shown in Algorithm~\ref{algo:rppo}. The implementation of RPPO requires only a few lines of code modifications based on PPO.

\begin{figure*}[ht]
    \centering
    \includegraphics[width=0.8\textwidth]{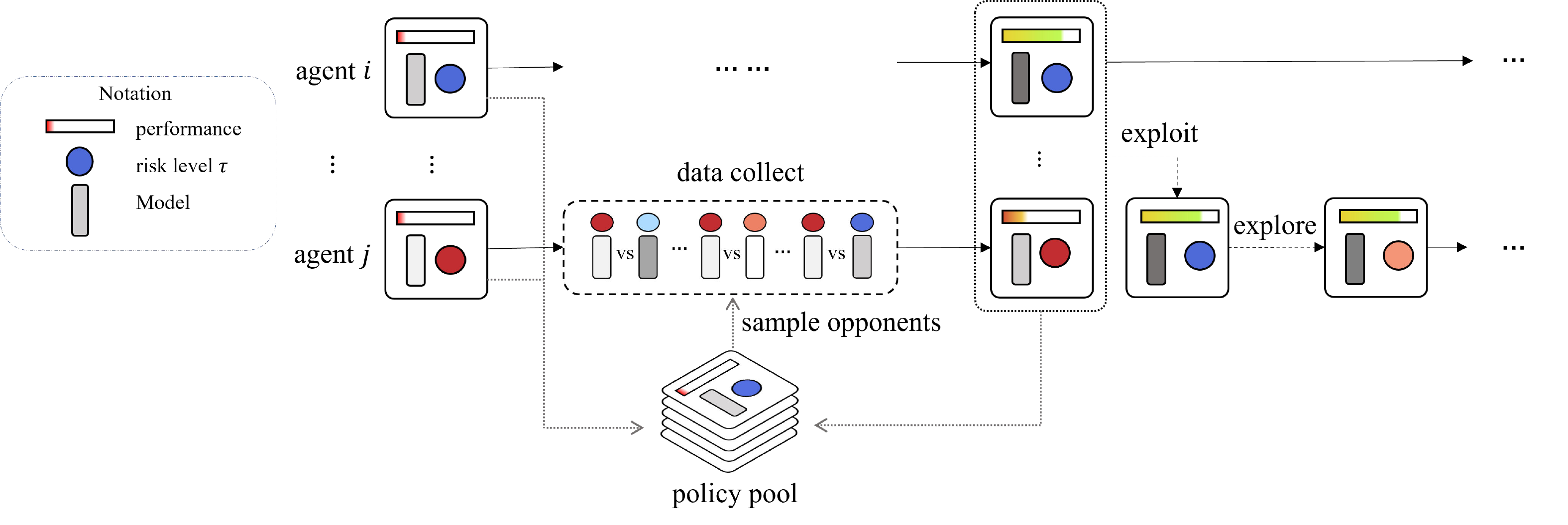}
    \caption{The framework of RPBT. We train a population of agents with different initialization of risk levels. During a training round, each agent spawns a number of subprocesses to collect data against randomly selected opponents from the policy pool. We use RPPO to update models under the specific risk level. The policies updated are added to the policy pool. If an agent in the population is under-performing, it will \emph{exploit} (copy) the model parameters and the risk level $\tau$ of a better-performing agent, and it will \emph{explore} a new risk level by adding a perturbation to the better-performing agent's risk level for the following training.}
    \label{fig:pbt_rppo}
\end{figure*}

\subsection{Toy example}
\label{sec:toy}
We use a grid world~\cite{deletang2021model} shown in Fig.~\ref{fig:toy.a} to illustrate the ability of RPPO to learn diverse risk preferences with different risk levels $\tau$. The grid world consists of an agent, a flag, water, and strong wind. The agent's goal is to reach the flag position within 25 steps while avoiding stepping into the water. When reaching the flag, the agent receives a +1 bonus and is done, but each step spent in the water results in a -1 penalty. In addition, the strong wind randomly pushes the agent in one direction with 50\% probability. In this world, risk (uncertainty) comes from the strong wind.

We train 9 RPPO agents with $\tau \in \{ 0.1, 0.2, ..., 0.9\}$ to investigate how risk level $\tau$ affects the behaviors of policies. We illustrate the state-visitation frequencies computed from the 1,000 rollouts, as shown in Fig.~\ref{fig:toy}. Three modes of policies emerge here: risk-averse ($\tau \in \{0.2,0.3,0.4\}$), taking the longest path away from the water; risk-neutral ($\tau \in \{0.5,0.6,0.7\}$), taking the middle path, and risk-seeking ($\tau \in \{0.8,0.9\}$), taking the shortest route along the water. Interestingly, agents with $\tau=0.1$ are too conservative and do not always reach the flag, which is consistent with risk-averse behavior styles.

\section{RPBT} \label{sec:pbt}

\begin{figure*}[!ht]
    \centering
    \includegraphics[width=0.85\textwidth]{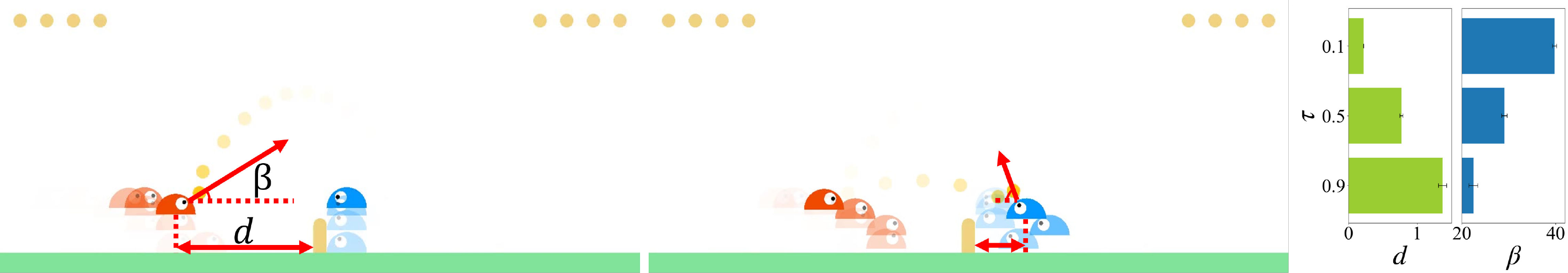}
    \caption{Illustration of diverse risk preferences in Slimevolley. Risk-seeking agent (left) stands farther from the fence and hit the ball at a lower angle while risk-averse agent (right) does the opposite.}
    \label{fig:volley_diverse}
\end{figure*}
\begin{figure*}[!ht]
    \centering
    \subfigure{}{
    \label{fig:ant_diverse_1}
    \includegraphics[width=0.12\textwidth]{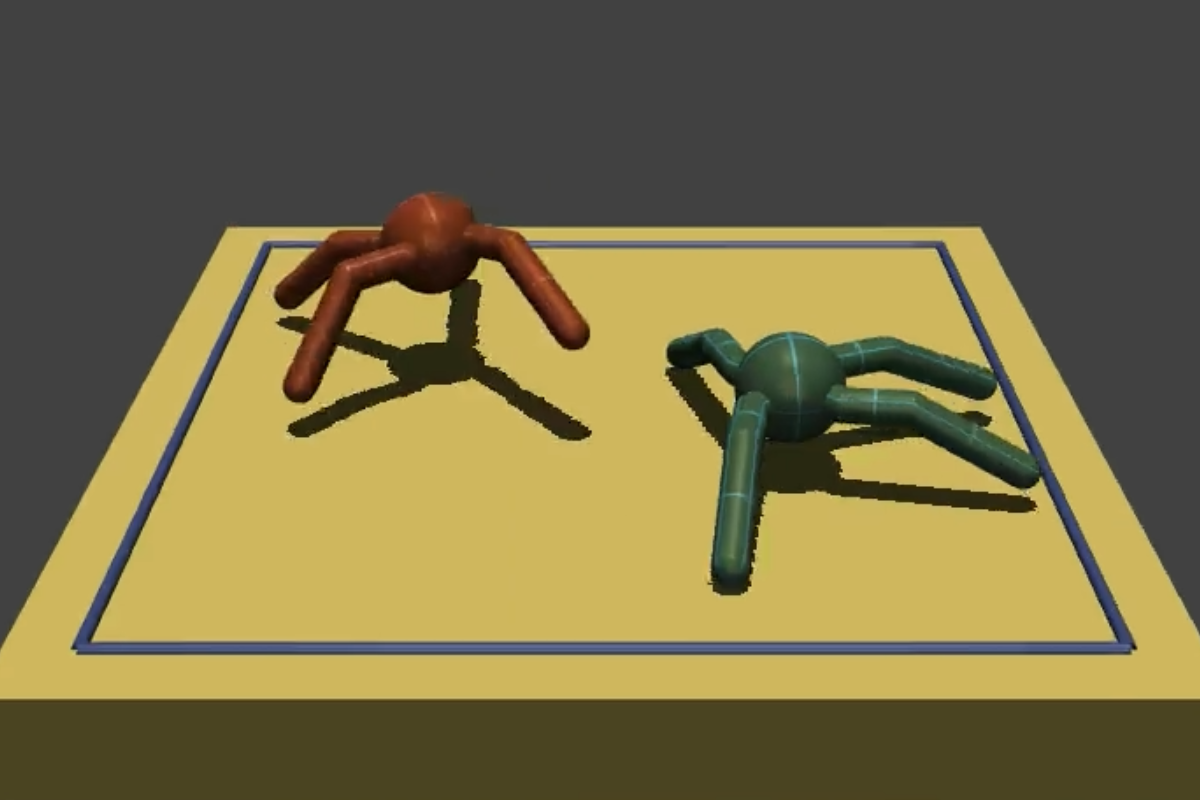}
    }
    \subfigure{}{
    \label{fig:ant_diverse_2}
    \includegraphics[width=0.12\textwidth]{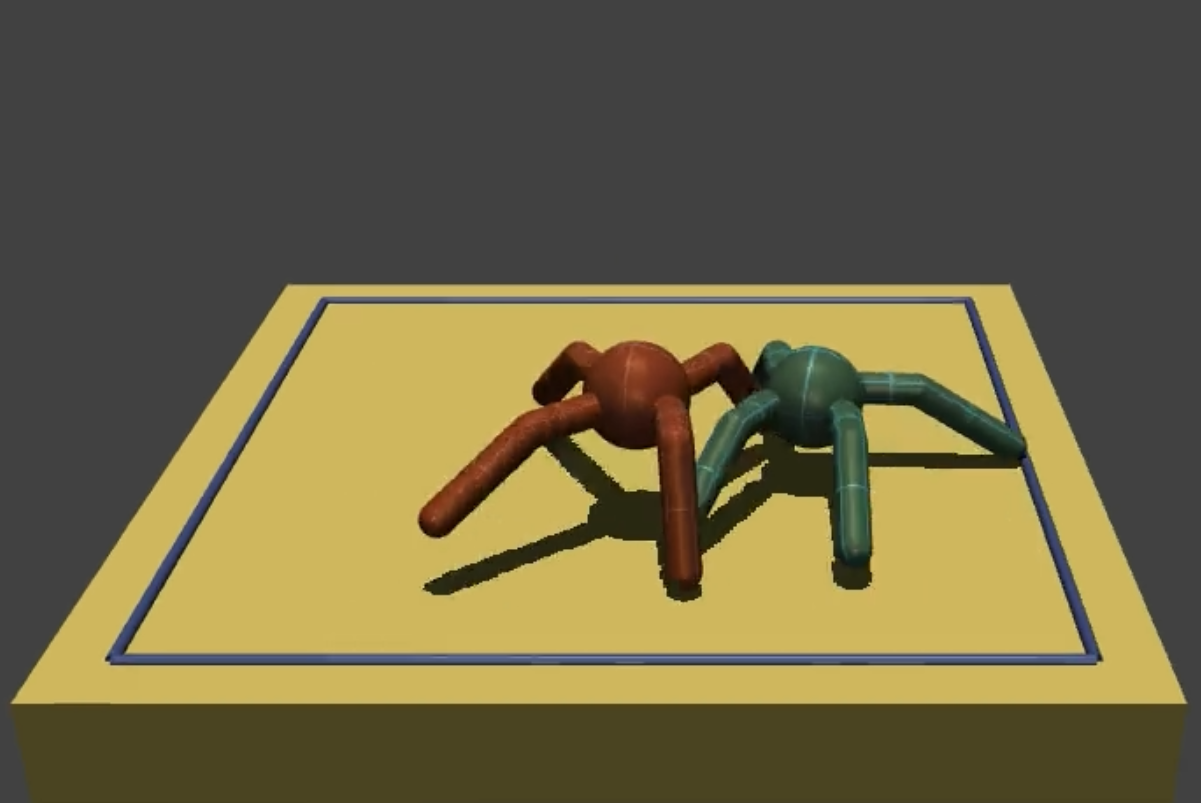}
    }
    \subfigure{}{
    \label{fig:ant_diverse_3}
    \includegraphics[width=0.12\textwidth]{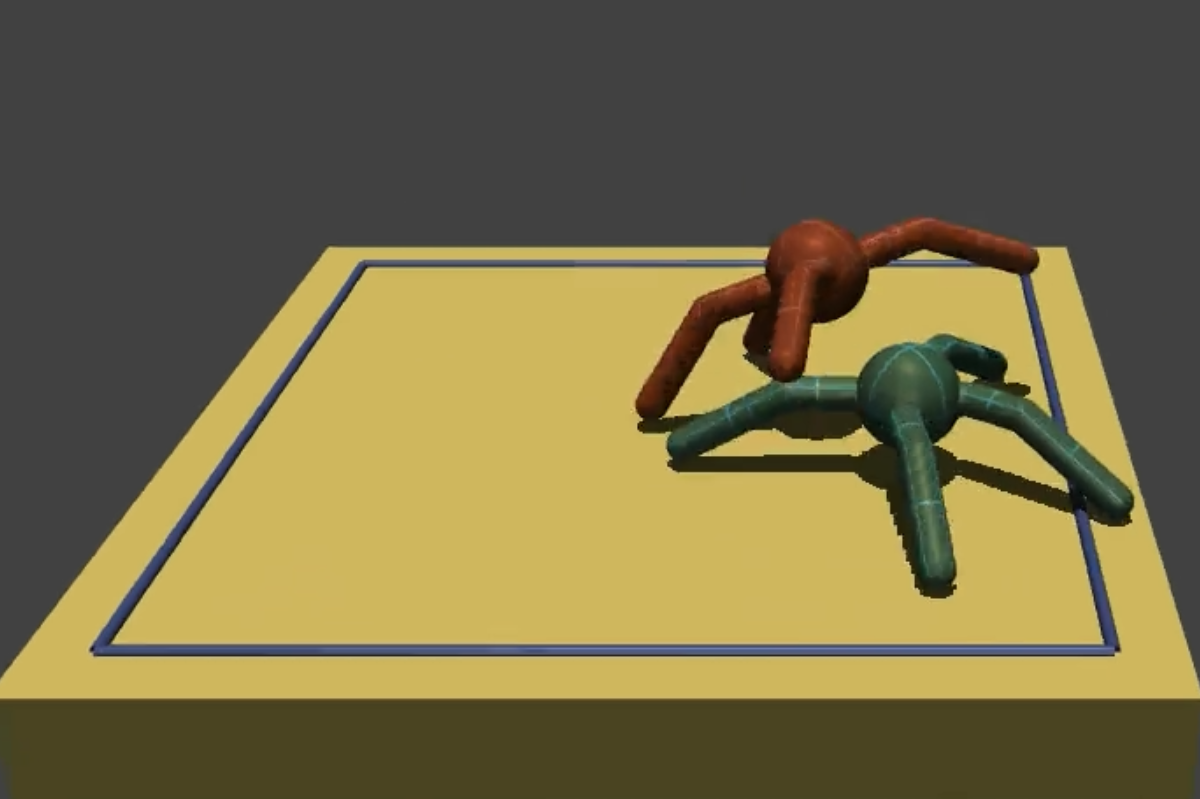}
    }
    \subfigure{}{
    \label{fig:ant_diverse_4}
    \includegraphics[width=0.12\textwidth]{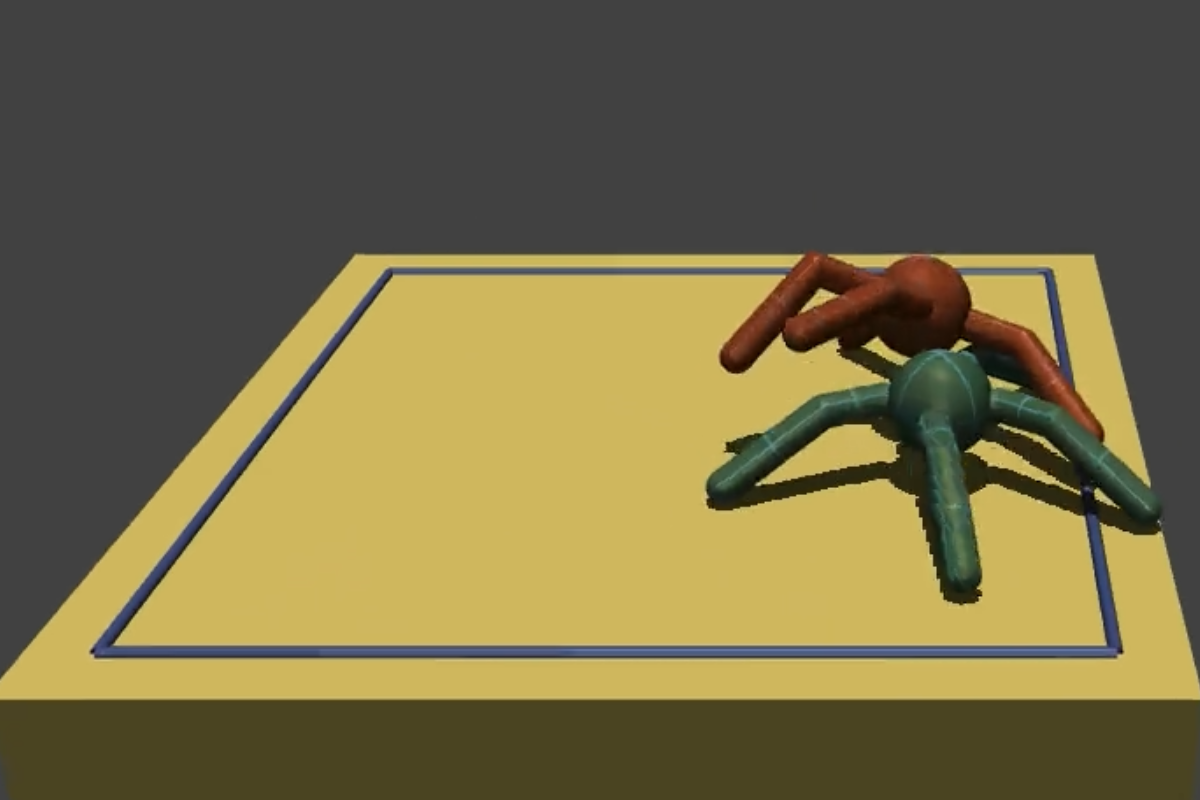}
    }
    \subfigure{}{
    \label{fig:ant_diverse_5}
    \includegraphics[width=0.12\textwidth]{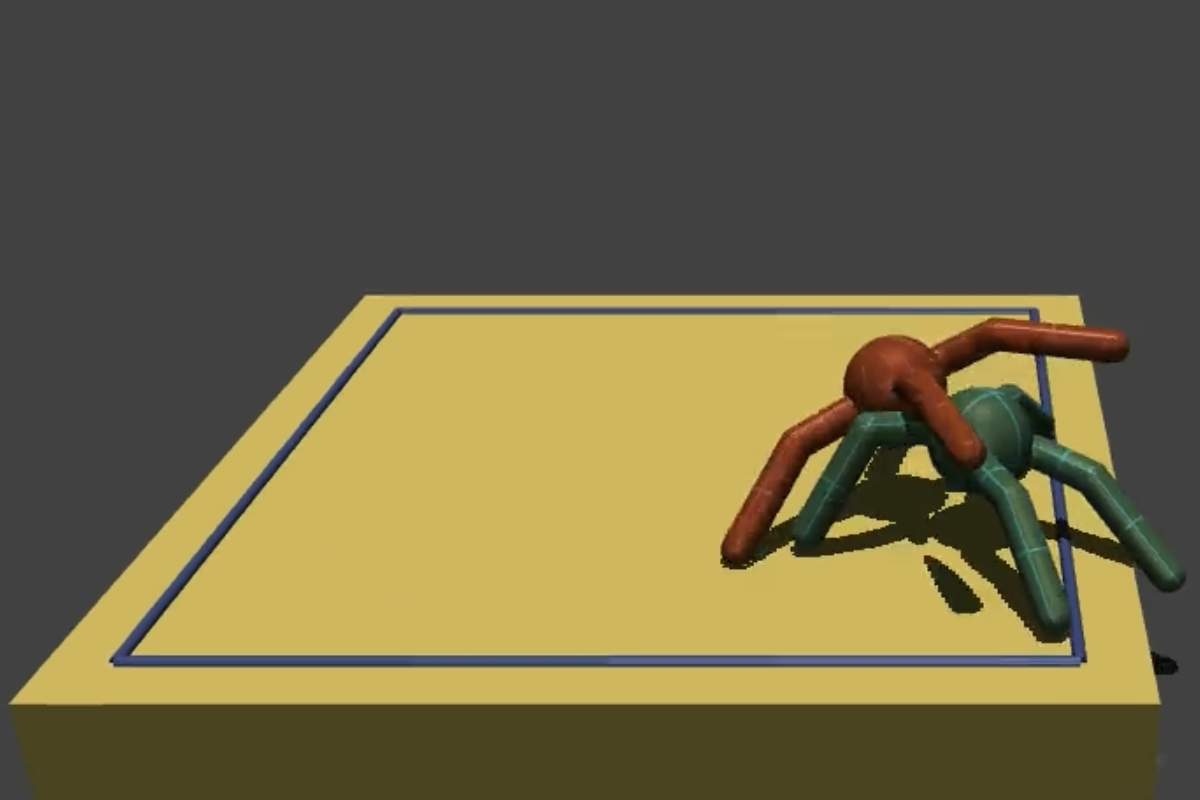}
    }
    \subfigure{}{
    \label{fig:ant_diverse_6}
    \includegraphics[width=0.12\textwidth]{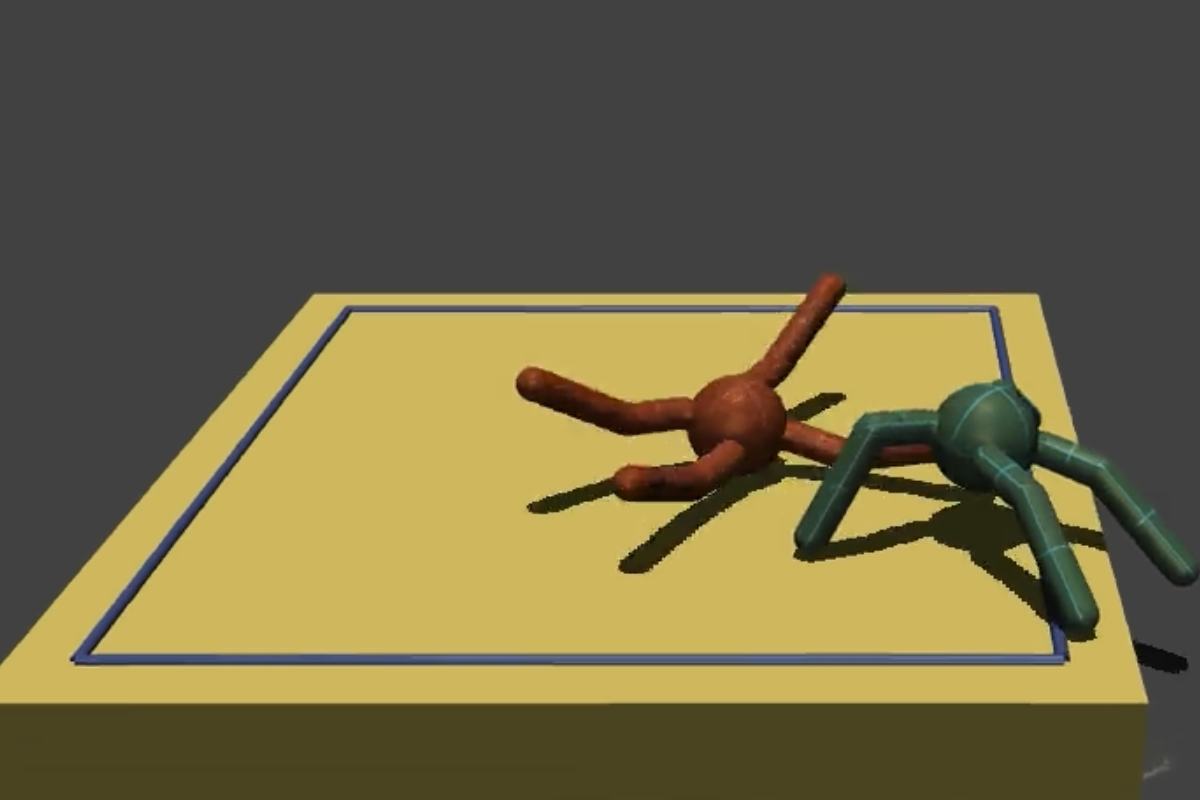}
    }
    \caption{Illustration of diverse risk preferences in Sumoants. The risk-seeking agent (red) constantly attacks, while the risk-averse agent (green) assumes a defensive stance.}
    \label{fig:ant_diverse}
\end{figure*}

In this section, we present our population-based self-play method based on RPPO, which we refer to as RPBT, as illustrated in Fig.~\ref{fig:pbt_rppo}. Our proposed RPBT stabilizes the learning process by concurrently training a diverse population of agents who learn by playing with each other. Moreover, different agents in the population have varying risk preferences in order to promote the learned policies to be diverse and not homogeneous. In contrast to previous work~\cite{parker-holder2020effective,lupu2021trajectory,zhao2022maximum}, RPBT does not include population-level diversity objectives. Instead, each agent acts and learns to maximize its individual rewards under the specific risk preference settings. This makes RPBT more easier to implement and scale.

In our RPBT, the value of the risk level $\tau \in (0,1)$ controls the agent’s risk preference. Since it is impossible to cover all possible risk levels within the population, we dynamically adjust the risk levels during training, particularly for those with poor performance. We adopt the exploitation and exploration steps in PBT~\cite{jaderberg2017population} to achieve this auto-tuning goal. In the exploitation step, a poorly performing agent can directly copy the model parameters and risk level $\tau$ from a well-performing agent to achieve equal performance. Then in the exploration step, the newly copied risk level is randomly perturbed by noise to produce a new risk level. In our practice, we introduce a simple technique:
\begin{itemize}
    \item \emph{Exploitation}. We rank all agents in the population according to ELO score, indicating performance. If an agent's ELO score falls below a certain threshold, it is considered an underperforming agent, and its model parameters and risk level  will be replaced by those from a superior agent.
    \item \emph{Exploration}. The risk level of the underperforming agent is further updated by adding a noise term varying between -0.2 and 0.2.
\end{itemize}
We find that this technique allows RPBT to explore almost all possible values of $\tau$ during training. At the later stage of training, the values of $\tau$ will converge to an interval with better performance while maintaining diversity and avoiding harmful values.

Additionally, we treat all agents of the population and their historical copies as a policy pool from which opponents are uniformly sampled for self-play. Actually, this approach of sampling opponents is commonly referred to as FSP~\cite{vinyals2019grandmaster}, which ensures that an agent must be able to defeat random old versions of any other agent in the population in order to continue learning adaptively~\cite{bansal2018emergent}. Since any approach of sampling opponents can be adapted to our method, we do not dwell on more complicated opponent sampling techniques~\cite{vinyals2019grandmaster, openai2018openai}. Moreover, we utilize FSP for all our experiments to ensure fair comparisons.

Fig.~\ref{fig:pbt_rppo} illustrated one training round of RPBT. Each training round consists of uniformly sampling opponents from the policy pool, collecting game experience, updating models using RPPO, and performing exploitation and exploration on risk levels. After training is ended, we select the agent with the highest ELO score from the population to serve as the evaluation agent.

\section{Experiments}
In our experiments, we aim to answer the following questions: \textbf{Q1}, can RPPO generate policies with diverse risk preferences in competitive games? \textbf{Q2}, how does RPBT perform compared to other methods in competitive games? Some results of the ablation experiments on RPBT are presented in the Appendix~\ref{app:additional}. All the experiments are conducted with one 64-core CPU and one GeForce RTX 3090 GPU.

\subsection{Environment setup}
\label{sec:envsetup}
We consider two competitive multi-agent benchmarks: \textbf{Slimevolley}~\cite{ha2020slime} and  \textbf{Sumoants}~\cite{al-shedivat2018continuous}. Slimevolley is a two-agent volleyball game where the action space is discrete. The goal of each agent is to land the ball on the opponent's field, causing the opponent to lose lives. Sumoants is a two-agent game based on MuJoCo where the action space is continuous. Two ants compete in a square area, aiming to knock the other agent to the ground or push it out of the ring. More details about the two benchmarks are given in Appendix~\ref{app:env}.

 \begin{table}[b]
    \centering
    \subtable[Slimevolley]{
    \resizebox{0.22\textwidth}{!}{
    \begin{tabular}{lcccc}
        \toprule
        \diagbox{Play A}{Play B} & RPBT        & PP     & SP \\
        \midrule
        RPBT(ours)               & -           & \textbf{56\%}    & \textbf{64\%}   \\
        PP                       & 44\%        & -                & 56\%   \\
        SP                       & 36\%        & 44\%      & -     \\
        \bottomrule
    \end{tabular}
    }}
    \subtable[Sumoants]{
    \resizebox{0.22\textwidth}{!}{
    \begin{tabular}{lcccc}
        \toprule
        \diagbox{Play A}{Play B}                   & RPBT       & PP     & SP \\
        \midrule
        RPBT(ours)               & -     & \textbf{63\%}    & \textbf{67\%}   \\
        PP                   & 37\%         & -       & 53\%    \\
        SP                  & 33\%         & 47\%      & -     \\
        \bottomrule
    \end{tabular}
    }}
    \caption{RPBT performing against basic baselines.}
    \label{tab:standardbaselines}
\end{table}

 \begin{table*}[!ht]
    \centering
    \subtable[Slimevolley]{
    \resizebox{0.42\textwidth}{!}{
    \begin{tabular}{lcccccc}
        \toprule
        \diagbox{Play A}{Play B}     & RPBT         & MEP     & TrajeDi & DvD  & RR & PSRO \\
        \midrule
        RPBT(ours)                         & -            & \textbf{64\%}    & \textbf{59\%}   & 48\% & \textbf{60\%} & \textbf{53\%}  \\
        MEP       & 36\%         & -                & 46\%            & 32\%          &  39\%         & 32\% \\
        TrajeDi & 41\%         & 53\%             & -               & 38\%          &  38\%         & 45\% \\
        DvD & 52\%      & 63\%             & 61\%            & -             &  58\%         & 63\%  \\
        RR        & 37\%      & 52\%             & 63\%            & 42\%          & -             & 54\%  \\
        PSRO    & 48\%         & 68\%             & 55\%            & 37\%          &  46\%         & - \\ 
        \bottomrule
    \end{tabular}
    }}
    \subtable[Sumoants]{
    \resizebox{0.42\textwidth}{!}{
    \begin{tabular}{lcccccc}
        \toprule
        \diagbox{Play A}{Play B}     & RPBT         & MEP     & TrajeDi & DvD  & RR & PSRO \\
        \midrule
        RPBT                         & -            & \textbf{58\%}    & \textbf{63\%}   & \textbf{54\%} & \textbf{56\%} & \textbf{53\%}  \\
        MEP                          & 42\%         & -                & 54\%            & 53\%          &  51\%         & 50\% \\
        TrajeDi                      & 37\%         & 46\%             & -               & 44\%          &  48\%         & 37\% \\
        DvD                          & 46\%      & 47\%             & 56\%            & -             &  52\%         & 43\%  \\
        RR                           & 44\%      & 49\%             & 52\%            & 48\%          & -             & 43\%  \\
        PSRO                         & 47\%         & 50\%             & 63\%            & 57\%          &  57\%         & - \\ 
         \bottomrule
    \end{tabular}
    }}
    \caption{RPBT performing against different methods.}
    \label{tab:baselines}
\end{table*}

\begin{figure*}[!h]
    \centering
    \includegraphics[width=0.80\textwidth]{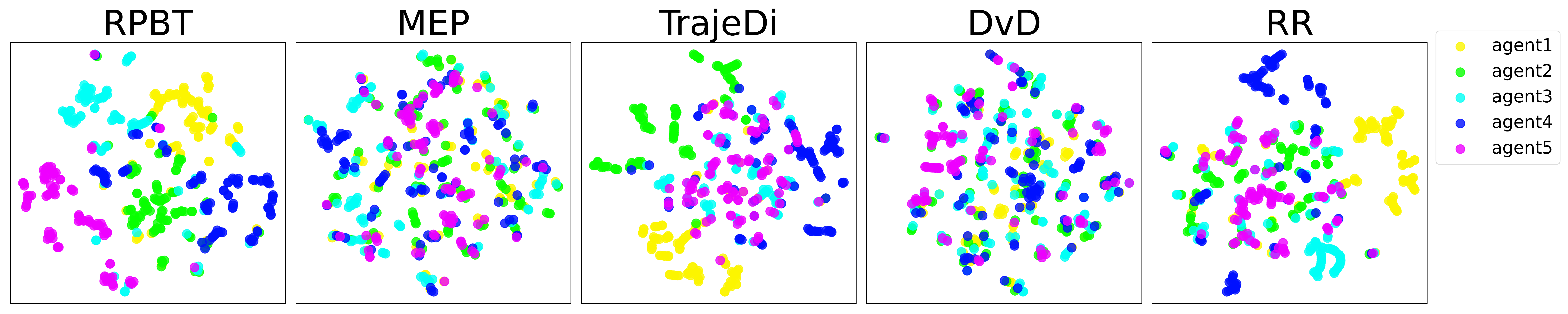}
    \caption{Comparing diversity of different methods with a population size of 5.}
    \label{fig:tsne}
\end{figure*}

\subsection{Diverse Risk Preferences Illustration} \label{sec:rppo_selfplay}



We trained RPPO agents via self-play, using various risk levels. As for Slimevolley, we pit the risk-seeking agent with $\tau=0.9$ against the risk-averse agent with $\tau=0.1$, as shown in Fig.~\ref{fig:volley_diverse}. We observe that the risk-seeking agent prefers to adjust its position to spike the ball so that the height of the ball is lower, while the risk-averse agent takes a more conservative strategy and just focuses on catching the ball. To further quantify this phenomenon, we pit the three agents $\tau \in \{0.1,0.5,0.9\}$ versus the $\tau=0.5$ agent and  compute two metrics average over 200 rollouts: the distance to the fences (denoted as $d$) and hitting angle (denoted as $\beta$). As $\tau$ increases, $\beta$ shows a monotonically decreasing trend, and $d$ shows a monotonically increasing trend. Because the physics engine of the volleyball game is a bit "dodgy", the agent can hit the ball with a low height only when standing far away from the fence.

As for Sumoants, a typical play between a risk-seeking agent with $\tau=0.7$ and a risk-averse agent with $\tau=0.3$ is illustrated in Fig.~\ref{fig:ant_diverse}. We observed that the risk-averse agent tends to maintain a defensive stance: four legs spread out, lowering its center of gravity and holding the floor tightly to keep it as still as possible. In contrast, the risk-seeking agent frequently attempts to attack. We also provide multiple videos for further observation on our code website.

\subsection{Results Compared with other methods} \label{sec:rppo_pbt}

We trained RPBT with population size 5 and set initial risk levels to $\{0.1, 0.4,0.5,0.6,0.9\}$ for all the experiments. To ensure fairness in our comparison, all the baselines were integrated with PPO and set to the same population size if necessary. See Appendix \ref{appendix:implement} for more implementation details. The baseline methods includes:
\begin{itemize}
    \item Basic baselines: self-play (\textbf{SP}), where agents learn solely through competing with themselves, and population-based self-play (\textbf{PP}), where a population of agents are co-trained through randomly competing against one another. 
    \item Population-based methods: \textbf{MEP}~\cite{zhao2022maximum}, \textbf{TrajeDi}~\cite{lupu2021trajectory}, \textbf{DvD}~\cite{parker-holder2020effective}, and \textbf{RR} (Reward Randomization, similar ideas with~\cite{tang2021discovering}). 
    \item Game-theoretic method: \textbf{PSRO}~\cite{lanctot2017unified}.
\end{itemize}

For each method, we trained 3 runs using different random seeds and selected the one with the highest ELO score for evaluation. Then, we calculated the win rate matrix between the methods, recording the average win rate corresponding to 200 plays. Tab.\ref{tab:standardbaselines} shows the results among RPBT and basic methods. We observe that RPBT exhibits superior performance compared to the baselines. Therefore, it is crucial to integrate diversity-oriented methods like RPBT in self-play. Tab.\ref{tab:baselines} shows the results among RPBT, population-based methods, and game-theoretic methods. We observed that RPBT outperforms MEP,
TrajeDi, and RR. Moreover, RPBT achieves performance comparable to that of DvD in Slimevolley but better in Sumoants. Furthermore, RPBT performs slightly better than PSRO. However, PSRO is more computationally intensive (While PSRO requires 34 hours for a single run, other methods only require almost 12 hours).  

In order to further compare the diversity in population-based methods, we let each agent in the population play against itself. We then extract the first 100 states from the game trajectories and use t-SNE~\cite{van2008visualizing} to reduce the states to 2 dimensions for visualizing. Fig.\ref{fig:tsne} shows the results. We observed that the trajectories of agents in the RPBT population are distinct from each other. In comparison, TrajeDi and RR exhibit a certain level of diversity, whereas MEP and DvD show no indications of diversity. Notably, these population-based methods require the introduction of an additional diversity objective at the population level. This causes higher complexity to implementation as well as hyperparameter tuning due to the trade-off between diversity and performance. Overall, RPBT is a favorable approach for both learning diverse behaviors and enhancing performance. 



\section{Conclusion}
Observing that learning to be diverse is essential for addressing the overfitting problem in self-play, we propose that diversity could be introduced in terms of risk preferences from different agents. Specifically, we propose a novel Risk-sensitive PPO (RPPO) approach to learn policies that align with desired risk preferences. Furthermore, we incorporate RPPO into a population-based self-play framework, which is easy to implement, thereby introducing the RPBT approach. In this approach, a population of agents, with diverse and dynamically adjusted risk preferences, compete with one another. We showed that our method can generate diverse modes of policies and achieve comparable or superior performance over existing methods. To the best of our knowledge, this is the first work that uses a risk-oriented approach to train a diverse population of agents. Future directions include developing a meta-policy that can adaptively select the optimal risk level according to the opponent's strategy and exploring the applicability of RPPO to safe reinforcement learning scenarios.

\newpage

\section{Acknowledgments}
This work was supported by National Natural Science Foundation of China under Grant No. 62192751, in part by the National Science and Technology Innovation 2030 - Major Project (Grant No. 2022ZD0208804), in part by Key R\&D Project of China under Grant No. 2017YFC0704100, the 111 International Collaboration Program of China under Grant No. BP2018006, and in part by the BNRist Program under Grant No. BNR2019TD01009, the National Innovation Center of High Speed Train R\&D project (CX/KJ-2020-0006), in part by the InnoHK Initiative, The Government of HKSAR; and in part by the Laboratory for AI-Powered Financial Technologies.




\bibliography{aaai24}

\onecolumn
\appendix
\section{Missing Proofs} \label{app:theory}
Our expectile Bellman operator shares similarities with the operator proposed in~\cite{ma2022offline}, which offers a smooth interpolation between the Bellman expectation operator and the optimality operator, ensuring appropriate conservatism in offline RL. However, their operator assumes deterministic dynamics, while ours does not, enabling us to use it for risk-sensitive learning where the risk arises from stochastic dynamics.
The proof of Proposition~\ref{proposition:one_contraction} and~\ref{proposition:one_monotoic}  shares same ideas with \cite{ma2022offline}.
\subsection{Proof of Proposition~\ref{proposition:one_contraction}}

\paragraph{Proposition 1}
For any $\tau \in(0,1), \mathcal{T}_\tau^\pi$ is a $\gamma_\tau$-contraction, where $\gamma_\tau=1-2 \alpha(1-\gamma) \min \{\tau, 1-\tau\}$.

\begin{proof}
We introduce two operators to simplify the proof:
\begin{align}
\mathcal{T}^{\pi}_{+}V(s) = V(s) + \mathbb{E}_{a}\mathbb{E}_{s^\prime}[\delta(s,a,s^\prime)]_{+}, \\
\mathcal{T}^{\pi}_{-}V(s) = V(s) + \mathbb{E}_{a}\mathbb{E}_{s^\prime}[\delta(s,a,s^\prime)]_{-}.
\end{align}

We first show that the two operators are non-expansion (i.e.  $||\mathcal{T}^{\pi}_{+}V_1-\mathcal{T}^{\pi}_{+}V_2)||_{\infty} \leq ||V_1 -V_2||_{\infty}$). For any $V_1, V_2$, we have
\begin{align}
\mathcal{T}^{\pi}_{+}V_1(s) - \mathcal{T}^{\pi}_{+}V_2(s) = \mathbb{E}_{a}\mathbb{E}_{s^\prime}[[\delta_1(s,a,s^\prime)]_{+}+V_1(s) - ([\delta_2(s,a,s^\prime)]_{+}+V_2(s))].
\end{align}

In terms of $([\delta_1(s,a,s^\prime)]_{+}+V_1(s) - ([\delta_2(s,a,s^\prime)]_{+}+V_2(s))$, we can discuss into four cases:
\begin{itemize}
\item $\delta_1 \geq 0, \delta_2 \geq 0$, then $[\delta_1(s,a,s^\prime)]_{+}+V_1(s) - ([\delta_2(s,a,s^\prime)]_{+}+V_2(s)) = \gamma[V_1(s^{\prime})-V_2(s^{\prime})]$.
\item $\delta_1 < 0, \delta_2 < 0$, then $[\delta_1(s,a,s^\prime)]_{+}+V_1(s) - ([\delta_2(s,a,s^\prime)]_{+}+V_2(s)) = V_1(s) - V_2(s)$.
\item $\delta_1 \geq 0, \delta_2 < 0$, then
\begin{equation}
  \begin{aligned}
  &[\delta_1(s,a,s^\prime)]_{+} +V_1(s) - ([\delta_2(s,a,s^\prime)]_{+} + V_2(s)) \\
  &= [r(s,a,s^{\prime}) + \gamma V_1(s^{\prime})] - V_2(s) \\
  & < [r(s,a,s^{\prime}) + \gamma V_1(s^{\prime})] - [r(s,a,s^{\prime}) + \gamma V_2(s^{\prime})] \\
  &= \gamma [V_1(s^{\prime})-V_2(s^{\prime})],
  \end{aligned}
\end{equation}
where the inequality comes from $\delta_2 < 0$.
\item $\delta_1 < 0, \delta_2 \geq 0$, then
\begin{equation}
  \begin{aligned}
  &[\delta_1(s,a,s^\prime)]_{+} +V_1(s) - ([\delta_2(s,a,s^\prime)]_{+} + V_2(s)) \\
  &= V_1(s) - [r(s,a,s^{\prime}) + \gamma V_2(s^{\prime})] \\
  & \leq V_1(s) - V_2(s),
  \end{aligned}
\end{equation}
 where the inequality comes from $\delta_2 \geq 0$.
\end{itemize}

Therefore, we have $\mathcal{T}^{\pi}_{+}V_1-\mathcal{T}^{\pi}_{+}V_2 \leq ||V_1-V_2||_{\infty}$. The proof for $\mathcal{T}^{\pi}_{-}$ is similar.  With $\mathcal{T}^{\pi}_{+}$ and $\mathcal{T}^{\pi}_{-}$, we can rewrite $\mathcal{T}^{\pi}_{\tau}$ as:
\begin{equation}\label{equ:operator_rewrite}
\begin{aligned}
\mathcal{T}_\tau^\pi V(s) &=V(s)+2 \alpha \mathbb{E}_{a} \mathbb{E}_{s^\prime}\left[\tau[\delta(s, a, s^\prime)]_{+}+(1-\tau)[\delta(s, a, s^\prime)]_{-}\right] \\
&=(1-2 \alpha) V(s)+2 \alpha \tau\left(V(s)+\mathbb{E}_{a}\mathbb{E}_{s^\prime}[\delta(s, a, s^\prime)]_{+}\right)+2 \alpha(1-\tau)\left(V(s)+\mathbb{E}_{a}\mathbb{E}_{s^\prime}[\delta(s, a, s^\prime)]_{-}\right) \\
&=(1-2 \alpha) V(s)+2 \alpha \pi \mathcal{T}_{+}^\pi V(s)+2 \alpha(1-\tau) \mathcal{T}_{-}^\pi V(s).
\end{aligned}
\end{equation}

If $\tau = 1 / 2$, we have
\begin{equation}
\begin{aligned}
\mathcal{T}_{1 / 2}^\pi V(s) =(1-2 \alpha) V(s)+\alpha\left(\mathcal{T}_{+}^\pi V(s)+\mathcal{T}_{-}^\pi V(s)\right).
\end{aligned}
\end{equation}

Then, we derive the contraction property of $\mathcal{T}^\pi_{1 / 2}$
\begin{equation}
\begin{aligned}
&\mathcal{T}_{1 / 2}^\pi V_1(s)-\mathcal{T}_{1 / 2}^\pi V_2(s) \\
&=V_1(s)+\alpha \mathbb{E}_{a}\mathbb{E}_{s^\prime}\left[\delta_1(s, a, s^\prime)\right]-\left(V_2(s)+\alpha \mathbb{E}_{a}\mathbb{E}_{s^\prime}\left[\delta_2(s, a, s^\prime)\right]\right) \\
&=(1-\alpha)\left(V_1(s)-V_2(s)\right)+\alpha \gamma \mathbb{E}_{a}\mathbb{E}_{s^\prime}\left[ V_1\left(s^{\prime}\right)- V_2\left(s^{\prime}\right)\right] \\
&\leq(1-\alpha)\left\|V_1-V_2\right\|_{\infty}+\alpha \gamma\left\|V_1-V_2\right\|_{\infty} \\
&=(1-\alpha(1-\gamma))\left\|V_1-V_2\right\|_{\infty} .
\end{aligned}
\end{equation}
Note that we recover the standard Bellman operator when $\alpha=1$.
Then, we consider the case of $\tau < 1/2$,
\begin{equation}
\begin{aligned}
&\mathcal{T}_\tau^\pi V_1(s)-\mathcal{T}_\tau^\pi V_2(s)\\
&=(1-2 \alpha)\left(V_1(s)-V_2(s)\right)+2 \alpha \tau\left(\mathcal{T}_{+}^\pi V_1(s)-\mathcal{T}_{+}^\pi V_2(s)\right)+
2 \alpha(1-\tau)\left(\mathcal{T}_{-}^\pi V_1(s)-\mathcal{T}_{-}^\pi V_2(s)\right)\\
&=(1-2 \alpha-2 \tau(1-2 \alpha))\left(V_1(s)-V_2(s)\right)+2 \tau\left(\mathcal{T}_{1 / 2}^\pi V_1(s)-\mathcal{T}_{1 / 2}^\pi V_2(s)\right)+
2 \alpha(1-2 \tau)\left(\mathcal{T}_{-}^\pi V_1(s)-\mathcal{T}_{-}^\pi V_2(s)\right)\\
&\leq(1-2 \alpha-2 \tau(1-2 \alpha))\left\|V_1-V_2\right\|_{\infty}+2 \tau(1-\alpha(1-\gamma))\left\|V_1-V_2\right\|_{\infty}+2 \alpha(1-2 \tau)\left\|V_1-V_2\right\|_{\infty}\\
&=(1-2 \alpha \tau(1-\gamma))\left\|V_1-V_2\right\|_{\infty}.
\end{aligned}
\end{equation}

Similarly, when $\tau>1 / 2$, we have $\mathcal{T}_\tau^\pi V_1(s)-\mathcal{T}_\tau^\pi V_2(s) \leq (1-2 \alpha(1-\tau)(1-\gamma)) || V_1-V_2 ||_{\infty} $. Gather them together, we have $\mathcal{T}_\tau^\pi V_1(s)-\mathcal{T}_\tau^\pi V_2(s) \leq (1-2 \alpha(1-\gamma) \min \{\tau, 1-\tau\}) || V_1-V_2 ||_{\infty} $. The proof is completed. 

We further discuss the value of step size $\alpha$. We want a larger step size in general, but $\alpha$ must satisfy that $V(s)+2\alpha\tau\delta(s,a, s') \leq \max\{r(s,a,s^{\prime})+\gamma V(s'), V(s)\}$ and $V(s) + 2\alpha(1-\tau)\delta(s,a, s')\ge \min\{r(s,a,s^{\prime})+\gamma V(s'), V(s)\}$, otherwise the $V$-value will be overestimated. Therefore, we can derive $\alpha \leq \frac{1}{2\max\{ \tau, 1-\tau \}}$, and we set $\alpha = \frac{1}{2\max \{ \tau, 1-\tau \}}$.
\end{proof}

\subsection{Proof of Proposition~\ref{proposition:one_monotoic}}
\paragraph{Proposition 2} Let $V_{\tau}^*$ denote the fixed point of $\mathcal{T}_\tau^\pi$. For any $\tau, \tau^{\prime} \in(0,1)$, if $\tau^{\prime} \geq \tau$, we have $V_{\tau^{\prime}}^*(s) \geq V_\tau^*(s), \forall s \in S$.

\begin{proof}
Based on equation~\ref{equ:operator_rewrite}, we have 
\begin{equation}
\begin{aligned}
&\mathcal{T}_{\tau^{\prime}}^\pi V(s)-\mathcal{T}_\tau^\pi V(s) \\
=&(1-2 \alpha) V(s)+2 \alpha \tau^{\prime} \mathcal{T}_{+}^\pi V(s)+2 \alpha\left(1-\tau^{\prime}\right) \mathcal{T}_{-}^\pi V(s) -  \\
& \quad \left((1-2 \alpha) V(s)+2 \alpha \tau \mathcal{T}_{+}^\pi V(s)+2 \alpha(1-\tau) \mathcal{T}_{-}^\pi V(s)\right) \\
=& 2 \alpha\left(\tau^{\prime}-\tau\right)\left(\mathcal{T}_{+}^\pi V(s)-\mathcal{T}_{-}^\pi V(s)\right) \\
=& 2 \alpha\left(\tau^{\prime}-\tau\right) \mathbb{E}_{a, s^{\prime}}\left[[\delta(s, a, s^\prime)]_{+}-[\delta(s, a, s^\prime)]_{-}\right] \geq 0,
\end{aligned}
\end{equation}
which means $\mathcal{T}_{\tau^{\prime}}^\mu \geq \mathcal{T}_\tau^\mu$ if $\tau^{\prime} \geq \tau$. 
Then we have $\mathcal{T}_{\tau^{\prime}}^\pi V_\tau^* \geq \mathcal{T}_{\tau}^\pi V_\tau^*$. 
Since $V_{\tau}^*$ is the fixed point of $\mathcal{T}_{\tau}^\pi$,
we have $\mathcal{T}_{\tau}^\pi V_{\tau}^*=V_{\tau}^*$. Thus, we obtain $V_{\tau}^*=\mathcal{T}_{\tau}^\pi V_\tau^* \leq \mathcal{T}_{\tau^{\prime}}^\pi V_{\tau}^*$. Repeatedly applying $\mathcal{T}_{\tau^{\prime}}^\pi$ and using its monotonicity, we have $V_{\tau}^* \leq \mathcal{T}_{\tau^{\prime}}^\pi V_{\tau}^* \leq\left(\mathcal{T}_{\tau^{\prime}}^\pi\right)^{\infty} V_\tau^*=V_{\tau^{\prime}}^*$.
\end{proof}

\subsection{Proof of Proposition~\ref{proposition:one_risk}}
\paragraph{Proposition 3} Let $V^*_{\tau}$, $V^*_{best}$, and $V^{*}_{worst}$ respectively denote the fixed point of expectile Bellman operator, best-case Bellman operator and worst-case Bellman operator. We have 
\begin{equation}
    V_{\tau}^*= 
    \begin{cases}
        V_{worst}^* & \text { if } \tau \rightarrow 0 \\ 
        V_{best}^* & \text { if } \tau \rightarrow 1.
    \end{cases}
\end{equation}

\begin{proof}
We give the proof of $\tau \rightarrow 0$, the case of $\tau \rightarrow 1$ is similar. We first show that $V^*_{worst}$ is also a fixed point of $\mathcal{T}_{-}^{\pi}$. 
Based on the definition of $\mathcal{T}_{worst}$, 
we have $V_{worst}^*=\min_{a,s^{\prime}}[R(s,a,s^{\prime})+\gamma V_{worst}^*(s^{\prime})]$, which infers that $\delta(s,a,s^\prime) \geq 0, \forall s \in \mathcal{S}, a \in A$. 
Therefore, we have $\mathcal{T}^{\pi}_{-}V_{worst}^*(s) = V_{worst}^*(s) + \mathbb{E}_{a \sim \pi}[\delta(s,a,s^\prime)]_{-} = V_{worst}^*(s)$. By setting $\tau \rightarrow 0$, we eliminate the item of $\mathcal{T}^{\pi}_{+}$. Further, we use the the contractive property of $\mathcal{T}_{\tau}^{\pi}$ to obtain the uniqueness of $V^{*}_{\tau}$.
\end{proof}

\subsection{Proof of Proposition~\ref{proposition:multi_contraction}}
\paragraph{Proposition 4} For any $\tau \in (0,1)$, $\mathcal{T}_{\tau, \lambda}^\pi$ is a $\gamma_{\tau,\lambda}$-contraction, where $\gamma_{\tau,\lambda} = \frac{(1 - \lambda)\gamma_\tau}{1 - \lambda \gamma_\tau}$.

\begin{proof}
We first show that for $h \in \mathbb{N}$, $(\mathcal{T}_\tau^\pi)^n$ is a $\gamma_\tau^n$-contraction. For any $V_1, V_2$, we have
\begin{equation}
\begin{aligned}
&(\mathcal{T}_\tau^\pi)^n V_1(s) -  (\mathcal{T}_\tau^\pi)^n V_2(s) \\
& = \mathcal{T}_\tau^\pi ((\mathcal{T}_\tau^\pi)^{n-1} V_1)(s) -
\mathcal{T}_\tau^\pi ((\mathcal{T}_\tau^\pi)^{n-1} V_2)(s) \\
& \leq \gamma_\tau \| (\mathcal{T}_\tau^\pi)^{n-1} V_1 - (\mathcal{T}_\tau^\pi)^{n-1} V_2 \|_\infty \\
& \leq \gamma_\tau^n \| V_1 - V_2 \|_\infty.
\end{aligned}
\end{equation}

Then, we have
\begin{equation}
\begin{aligned}
&\mathcal{T}_{\tau,\lambda}^{\pi}V_1(s) -  \mathcal{T}_{\tau,\lambda}^{\pi}V_2(s) \\
& = (1 - \lambda) \sum_{n=1}^{\infty} \lambda^{n-1} (\mathcal{T}_\tau^\pi)^{n} V_1(s) - (1 - \lambda) \sum_{n=1}^{\infty} \lambda^{n-1} (\mathcal{T}_\tau^\pi)^{n} V_2(s) \\
& \leq (1 - \lambda) \sum_{n=1}^{\infty} \lambda^{n-1} \gamma_\tau^n \| V_1 - V_2 \|_\infty \\
& \leq \frac{(1 - \lambda)\gamma_\tau}{1 - \lambda\gamma_\tau} \| V_1 - V_2 \|_\infty.
\end{aligned}
\end{equation}
\end{proof}

\subsection{Proof of Proposition~\ref{proposition:multi_monotoic}}

\paragraph{Proposition 5} Let $V_{\tau,\lambda}^*$ denote the fixed point of $\mathcal{T}_{\tau,\lambda}^\pi$. For any $\tau, \tau^{\prime} \in(0,1)$, if $\tau^{\prime} \geq \tau$, we have $V_{\tau^{\prime},\lambda}^*(s) \geq V_{\tau,\lambda}^*(s), \forall s \in S$.

\begin{proof}
Based on Proposition~\ref{proposition:one_monotoic}, we have $\mathcal{T}_{\tau^{\prime}}^\pi \geq \mathcal{T}_\tau^\pi$ , by recursively applying this, we have:
\begin{equation}
\begin{aligned}
&\mathcal{T}_{\tau^{\prime},\lambda}^{\pi}V(s) - \mathcal{T}_{\tau,\lambda}^{\pi}V(s) \\
&= (1-\lambda)\sum_{n=1}^{\infty}\lambda^{n-1}\left[(\mathcal{T}^{\pi}_{\tau^{\prime}})^{n}V(s)-(\mathcal{T}^{\pi}_{\tau})^{n}V(s)\right] \\
&=(1-\lambda) \sum_{n=1}^{\infty} \lambda^{n-1}\left[\mathcal{T}_{\tau^{\prime}}^\pi\left(\mathcal{T}_{\tau^{\prime}}^\pi\right)^{n-1} V(s)-\left(\mathcal{T}_\tau^\pi\right)^n V(s)\right] \\
&\geqslant(1-\lambda) \sum_{n=1}^{\infty} \lambda^{n-1}\left[\mathcal{T}_\tau^\pi\left(\mathcal{T}_{\tau^{\prime}}^\pi\right)^{n-1} V(s)-\left(\mathcal{T}_\tau^\pi\right)^n V(s)\right] \\
&\geqslant(1-\lambda) \sum_{n=1}^{\infty} \lambda^{n-1}\left[\left(\mathcal{T}_\tau^\pi\right)^2\left(\mathcal{T}_{\tau^{\prime}}^\pi\right)^{n-2} V(s)-\left(\mathcal{T}_\tau^\pi\right)^n V(s)\right] \\
&\quad \quad \quad \quad \quad\vdots \\
&\geqslant(1-\lambda) \sum_{n=1}^{\infty} \lambda^{n-1}\left[\left(\mathcal{T}_\tau^\pi\right)^n V(s)-\left(\mathcal{T}_\tau^\pi\right)^n V(s)\right] \\
&=0,
\end{aligned}
\end{equation}
which means $\mathcal{T}_{\tau^{\prime},\lambda}^{\pi}V(s) \geq \mathcal{T}_{\tau,\lambda}^{\pi}V(s) $ if $\tau^{\prime} \geq \tau$. Then, we have $\mathcal{T}_{\tau^{\prime},\lambda}^\pi V_{\tau,\lambda}^* \geq \mathcal{T}_{\tau,\lambda}^\pi V_{\tau,\lambda}^*$. 
Since $V_{\tau,\lambda}^*$ is the fixed point of $\mathcal{T}_{\tau,\lambda}^\pi$, we have $\mathcal{T}_{\tau,\lambda}^\pi V_{\tau,\lambda}^*=V_{\tau,\lambda}^*$.
Thus, we obtain $V_{\tau,\lambda}^*=\mathcal{T}_{\tau,\lambda}^\pi V_{\tau,\lambda}^* \leq \mathcal{T}_{\tau^{\prime},\lambda}^\pi V_{\tau,\lambda}^*$. 
Repeatedly applying $\mathcal{T}_{\tau^{\prime},\lambda}^\pi$ and using its monotonicity, we have $V_{\tau,\lambda}^* \leq \mathcal{T}_{\tau^{\prime},\lambda}^\pi V_{\tau,\lambda}^* \leq\left(\mathcal{T}_{\tau^{\prime},\lambda}^\pi\right)^{\infty} V_{\tau,\lambda}^*=V_{\tau^{\prime},\lambda}^*$.
\end{proof}

\subsection{Proof of Proposition~\ref{proposition:multi_risk}}
\paragraph{Proposition 6}
Let $V^*_{\tau,\lambda}$ as the fixed point of $\mathcal{T}^{\pi}_{\tau,\lambda}$, we have
\begin{equation}
    \lim _{\tau} V_{\tau,\lambda}^*= 
    \begin{cases}
        V_{worst}^* & \text { if } \tau \rightarrow 0 \\ 
        V_{best}^* & \text { if } \tau \rightarrow 1
    \end{cases}.
\end{equation}

\begin{proof}
Let $V^*_{\tau}$ denote the fixed point of of $\mathcal{T}^{\pi}_{\tau}$.
As for $\mathcal{T}^{\pi}_{\tau, \lambda}$, we have:
\begin{equation}
\begin{aligned}
V_{\tau, \lambda}^* &=(1-\lambda) \sum_{n=1}^{\infty} \lambda^{n-1}\left(\mathcal{T}_\tau^\pi\right)^n V_\tau^*(s) \\
&=(1-\lambda) \sum_{n=1}^{\infty} \lambda^{n-1} V_\tau^*(s) \\
&=V_\tau^*(s)
\end{aligned}
\end{equation}
\end{proof}
Proof is completed by making use of Proposition~\ref{proposition:one_risk}.

\section{Multi-step Expectile Bellman Operator}
We first demonstrate the relationship between the standard multi-step Bellman operator and GAE, and then explain why GAE cannot be directly applied to the multi-step expectile Bellman operator. Finally, we show how to compute the multi-step expectile Bellman operator in practice.

A sample form standard Bellman operator is defined as follows:
\begin{equation}
    \hat{\mathcal{T}}^{\pi} V(s_t) = r_t + \gamma V(s_{t+1}).
\end{equation}
The two-step Bellman operator can be derived as:
\begin{equation}
    (\hat{\mathcal{T}}^{\pi})^2V(s_t) = r_t + \gamma \hat{\mathcal{T}}^{\pi}V(s_{t+1}) = r_t + \gamma r_{t+1}+ \gamma^2V(s_{t+2}).
\end{equation}
And so on, we can derive the general formula of n-step Bellman operator:
\begin{equation}
    (\hat{\mathcal{T}}^{\pi})^nV(s_t) = \sum_{l=0}^{n-1}\gamma^{l} r_{t+l} + \gamma^nV(s_{t+n}).
\end{equation}
As the standard multi-step Bellman operator is an exponentially weighted sum of one-step to $\infty$-step, we can derive its relation to GAE:
\begin{equation}
\begin{aligned}
& \hat{\mathcal{T}}_{\lambda}^{\pi}V(s_t)=(1-\lambda) \sum_{n=1}^{\infty} \lambda^{n-1}\left(\hat{\mathcal{T}}^{\pi}\right)^n V\left(s_t\right) \\
& =(1-\lambda) \sum_{n=1}^{\infty} \lambda^{n-1}\left(\sum_{l=0}^{n-1} \gamma^l r_{t+l}+\gamma^n V\left(s_{t+n}\right)\right) \\
& =\sum_{n=1}^{\infty} \sum_{l=0}^{n-1}(1-\lambda) \lambda^{n-1} \gamma^l r_{t+1}+\sum_{n=1}^{\infty}(1-\lambda) \lambda^{n-1} \gamma^n  V (s_{t+n}) \\
& =\sum_{l=0}^{\infty} \sum_{n=l+1}^{\infty}(1-\lambda) \lambda^{n-1} \gamma^l r_{t+l}+\sum_{l=0}^{\infty}(1-\lambda) \lambda^l \gamma^{l+1} \left(s_{t+l+1}\right) \\
& =\sum_{l=0}^{\infty}(\lambda \gamma)^l \sum_{n=t+1}^{\infty}(1-\lambda) \lambda^{n-l-1} r_{t+l}+\sum_{l=0}^{\infty}(\lambda \gamma)^{l} \gamma V\left(s_{t+l+1}\right)-\sum_{l=0}^{\infty}(\lambda r)^{l+1} V\left(s_{t+l+1}\right) \\
& =\sum_{l=0}^{\infty}(\lambda \gamma)^l \left[r_{t+l}+\gamma V(s_{t+l+1}) \right]-\sum_{l=0}^{\infty}(\lambda \gamma)^{l+1} V\left(s_{t+l+1}\right) \\
& =\sum_{l=0}^{\infty}(\lambda r)^l\left[r_{t+l}+\gamma V\left(s_{t+l+1}\right)\right]-\sum_{l=0}^{\infty}(\lambda \gamma)^l V\left(s_{t+l}\right)+V\left(s_t\right) \\
& =\sum_{l=0}^{\infty}(\lambda r)^l\left[r_{t+1}+\gamma V\left(s_{t+l+1}\right)-V\left(s_{t+l}\right)\right]+V\left(s_t\right) \\
& =\sum_{l=0}^{\infty}(\lambda r)^l \delta_{t+l}^V+V\left(s_t\right) . \\
&
\end{aligned}
\end{equation}
This establishes that standard multi-step Bellman operator is equivalent to GAE plus the value function itself, which makes it easy for us to compute in practice.

However, when it comes to the multi-step Bellman expectile operator, this relationship no longer holds. Taking two-step operator for example, we can only derive to this point:
\begin{equation}
\left(\hat{\mathcal{T}}_\tau^\pi\right)^2 V\left(s_t\right)=\hat{\mathcal{T}}_\tau^\pi\left(V\left(s_t\right)+2 \alpha\left[\tau\left[r_t+ \gamma V\left(s_{t+1}\right)-V\left(s_t\right)\right]_{+} + (1-\tau) \left[r_t+\gamma V\left(s_{t+1}\right)-V\left(s_t\right)\right]_{-}\right]\right)
\end{equation}
Due to the nonlinear operations $[\cdot]_{+}$ and $[\cdot]_{-}$, it is no longer possible to derive the general formula of the n-step operator, and only the recurrence relation between the n-step operator and the (n-1)-step operator can be determined.

Therefore, in practice, we need to maintain a table, as shown in Tab.~\ref{tab:gae}, to record the operator value for each step. The nonzero values of the table form a triangle because of the terminated state. Let $d_t$ denotes the time length between $s_t$ and the terminated state, after normalizing the exponential coefficients to sum to one, we can derive the value of multi-step operator:

\begin{equation}
    \hat{\mathcal{T}}^{\pi}_{\tau,\lambda}V(s_t) = \frac{1-\lambda}{1-\lambda^{d_t}}\sum_{h=1}^{d_t} \lambda^{h-1} (\hat{\mathcal{T}}^{\pi}_{\tau,\lambda})^h V(s_t).
\end{equation}

The disadvantage of keeping such a table is that it increases storage and computation complexity. However, as $\lambda^n$ approaches 0 for a large $n$, we can alleviate this problem by restricting the amount of rows $n$ in the table, i.e. $n=\min \{50, T_{\max}\}$, where $T_{\max}$ denotes the maximum possible length of an episode. Empirically, training a RPPO agent takes 10 hours with the task of training 1e8 steps via self-play in Sumoants, while training a PPO agent requires 7 hours. We believe that this level of computational complexity is acceptable.

\begin{table}
    \centering
    \resizebox{\textwidth}{!}{
    \begin{tabular}{|c|c|c|c|c|c|c|c|c|}
    \hline
        Coefficient & $s_0$ & $s_1$ & $s_2$ & $\dots$ & $s_{T-3}$ & $s_{T-2}$ & $s_{T-1}$ & $s_{T}$ \\ \hline
        1 & $\hat{\mathcal{T}}^{\pi}_{\tau}V(s_0)$ & $\hat{\mathcal{T}}^{\pi}_{\tau}V(s_1)$ & $\hat{\mathcal{T}}^{\pi}_{\tau}V(s_2)$ & $\vdots$ & $\hat{\mathcal{T}}^{\pi}_{\tau}V(s_{T-3})$ & $\hat{\mathcal{T}}^{\pi}_{\tau}V(s_{T-2})$ & $\hat{\mathcal{T}}^{\pi}_{\tau}V(s_{T-1})$ &  ~  \\ \hline
        $\lambda$ & $(\hat{\mathcal{T}}^{\pi}_{\tau})^2V(s_0)$ & $(\hat{\mathcal{T}}^{\pi}_{\tau})^2V(s_1)$ & $(\hat{\mathcal{T}}^{\pi}_{\tau})^2V(s_2)$ & $\vdots$ & $(\hat{\mathcal{T}}^{\pi}_{\tau})^2V(s_{T-3})$ & $(\hat{\mathcal{T}}^{\pi}_{\tau})^2V(s_{T-2})$ & 0 & ~ \\ \hline
        $\lambda^2$ & $(\hat{\mathcal{T}}^{\pi}_{\tau})^3V(s_0)$ & $(\hat{\mathcal{T}}^{\pi}_{\tau})^3V(s_1)$ & $(\hat{\mathcal{T}}^{\pi}_{\tau})^3V(s_2)$ & $\vdots$ & $(\hat{\mathcal{T}}^{\pi}_{\tau})^3V(s_{T-3})$ & 0 & 0 & ~ \\ \hline
        $\vdots$ & $\vdots$ & $\vdots$ & $\vdots$ & $\vdots$ & $\vdots$ & $\vdots$ & $\vdots$ &  ~\\ \hline
        $\vdots$ & $\vdots$ & $\vdots$ & $\vdots$ & $\vdots$ & $\vdots$ & $\vdots$ & $\vdots$ &  ~\\ \hline
        $\lambda_{T-3}$ & $(\hat{\mathcal{T}}^{\pi}_{\tau})^{T-2}V(s_0)$ & $(\hat{\mathcal{T}}^{\pi}_{\tau})^TV(s_1)$ & $(\hat{\mathcal{T}}^{\pi}_{\tau})^{T-2}V(s_2)$ & $\vdots$ & 0 & 0 & 0 & ~\\ \hline
        $\lambda^{T-2}$ & $(\hat{\mathcal{T}}^{\pi}_{\tau})^{T-1}V(s_0)$ & $(\hat{\mathcal{T}}^{\pi}_{\tau})^{T-1}V(s_1)$ & 0 & $\vdots$ & 0 & 0 & 0 & ~ \\ \hline
        $\lambda^{T-1}$ & $(\hat{\mathcal{T}}^{\pi}_{\tau})^TV(s_0)$ & 0 & 0 & $\vdots$ & 0 & 0 & 0 & ~ \\ \hline
        Exponential sum & $\hat{\mathcal{T}}^{\pi}_{\tau,\lambda}V(s_0)$ & $\hat{\mathcal{T}}^{\pi}_{\tau,\lambda}V(s_1)$ & $\hat{\mathcal{T}}^{\pi}_{\tau,\lambda}V(s_2)$ & $\vdots$ & $\hat{\mathcal{T}}^{\pi}_{\tau,\lambda}V(s_{T-3})$ & $\hat{\mathcal{T}}^{\pi}_{\tau,\lambda}V(s_{T-2})$ & $\hat{\mathcal{T}}^{\pi}_{\tau,\lambda}V(s_{T-1})$ &  ~ \\ \hline
    \end{tabular}}
    \caption{The table that needs to be kept for computing the multi-step Expectile Bellman operator $\hat{\mathcal{T}}^{\pi}_{\tau,\lambda}$ in practice. Assuming we have collected one episode data, and $s_{T}$ is the terminated state.}
    \label{tab:gae}
\end{table}

\section{Environment Setup} \label{app:env}
\paragraph{Slimevolley}
Slimevolley~\cite{ha2020slime} is a two-agent competitive volleyball game where the action space is discrete. The goal of each agent is to land the ball on the opponent's field, causing the opponent to lose one life. When the opponent loses a life, the agent gains +1 bonus. If the agent loses a life, -1 penalty will be given. Each agent starts with five lives, and the game ends when any agent loses all the five lives or after 3000 time steps. As the game ends, the agent with more lives wins.

\paragraph{SumoAnts}
SumoAnts~\cite{al-shedivat2018continuous} is a two-agent competitive game based on MuJoCo where the action space is continuous. Two agents (ants) compete in a square area, aiming to knock the other agent to the ground or push it out of the ring. The winner receives +2000 reward, and the loser gets -2000 reward. If there is a draw, both agents gets -1000 reward. In addition, there are also dense exploration rewards, which is used to promote training. We remain consistent with the original paper.

\section{Implementation Details}\label{appendix:implement}

We provide a single file implementation of RPPO and a lightweight, scalable implementation of RPBT based on Ray~\cite{moritz2018ray}, which refers to CleanRL~\cite{huang2022cleanrl} and  MAPPO~\cite{yu2022surprising} respectively.

\subsection{Toy Example}
The experiment of toy example based on our single file implementation of RPPO. The network of RPPO agents consists of 2 linear layers with 128 units and a logit
layer with four actions(4 possible walking directions). The discount
factor $\gamma$ is set to 0.95, and $\lambda$ for computing advantages is 0.95. The inputs to the network are the one-hot encoding of the number of grids. Each agent was trained for 1000K steps with a batch size of 200, using Adam optimizer with learning rate 1e-4.

\subsection{Slimevolley \& Sumoants}
We use PPO with policy and value networks as the algorithm backbone of RPPO in all experiments. Both policy and value networks consist of a multi-layer perceptrons (MLP) with 2 linear layers, a GRU layer with same hidden size of 128. As the action space for Slimevolley is discrete, and for Sumoants is continuous. We use Categorical logits for Slimevolley and Gaussian logits for Sumoants. Specially for Gaussian logits, the mean of Gaussian policy is the output of the networks and the entries of a diagonal covariance matrix are also as trainable parameters. The policy outputs are clipped to lie within the control range. The PPO hyperparameters we use for each experiment is shown in Tab.\ref{tab:hyperparameters}. 
\begin{table}
    \centering
    \begin{tabular}{ccc}
        \toprule
        Hyperparamters    &  Slimevolley & Sumoants  \\
       \midrule
       Total training steps & 1e8 & 1e8 \\
       Hidden size & 128  & 128   \\
       Learning rate & 3e-4 & 3e-4 \\
       Batch size &  96000 & 96000  \\
       Minibatch size & 24000 & 24000 \\
        Optimizer  &  Adam    & Adam \\
        Learning rate &  3e-4  & 3e-4 \\
        Discount factor $\gamma$  & 0.995 & 0.995 \\
        Total training steps & 1e8  & 1e8  \\
        GRU horizontal length & 8 & 8 \\
        PPO update epochs & 4 & 4\\ 
        PPO clipping parameter & 0.2 & 0.2 \\
        Bonus entropy coefficient & 0.01 & 0.01 \\
        \bottomrule
    \end{tabular}
    \caption{Hyperparameter sheet of PPO}
    \label{tab:hyperparameters}
\end{table}

For all the experiments, we use population size 5 for RPBT and set the initial risk level hyperparameter $\tau$ as $\{0.1,0.4,0.5,0.6,0.9\}$. We use ELO score to measure the performance of the player in the population. An agent is under-performing when the difference of ELO
between it and a well-performing agent is greater than a certain threshold (250 for Sumoants, 500 for Slimevolley). We set ELO update interval to one training round. Let A, B denote the two players respectively. We start from the perspective of player A, the update process of ELO is divided into three steps:
\begin{itemize}
    \item Before the match, compute the expected score of A:
        \begin{equation}
        E_A=\frac{1}{1+10^{\left(R_B-R_A\right) / 400}},
        \end{equation}
        where $E_A$ is the expected score, $R_A$, $R_B$ refers to current ELO of A and B respectively, 
    \item After the match, computed the actual score $S_A$: Generally Win=1, Lose=0, Tie=0.5.
    \item Update ELO
        \begin{equation}
        R_A \leftarrow R_A+K\left(S_A-E_A\right),
        \end{equation}
        where $K$ can be considered as the learning rate.
\end{itemize}
We set the initial ELO score to 1000, and use $K=32$ for all the experiments. 

\subsection{Baselines}
All baseline methods are integrated with PPO and utilize the same hyperparameters as presented in Tab.\ref{tab:hyperparameters}. More over, all baselines except SP and PSRO set the population size to 5.

\paragraph{SP} Agents trained by self-play, there is no population.

\paragraph{PP} Agents trained by population-based self-play, each agent has a different random seed. 

\paragraph{DvD} DvD~\cite{parker-holder2020effective} utilizes the determinant of the kernel matrix of action embedding as an auxiliary loss. We use RPBF kernel matrix and embedding mini-batch size 512. The coefficient of DvD loss is set to 0.01 after searching in \{0.01, 0.1, 0.5\}.

\paragraph{TrajeDi} TrajeDi~\cite{lupu2021trajectory} utilizes the approximate Jensen-Shannon divergence with action discounting kernel as an auxiliary loss. We set the coefficient of trajectory diversity loss to 0.1 and action discount factor to 0.5.

\paragraph{MEP} MEP~\cite{zhao2022maximum} maximizes a lower bound of the average KL divergence and then assigns each agent an auxiliary reward to achieve diversity. The auxiliary reward form in each time step $t$ is $\tilde {r}=r\left(s_t, a_t\right)-\alpha \log \left(\bar{\pi}\left(a_t \mid s_t\right)\right)$, where $\bar{\pi}\left(a_t \mid s_t\right)$ refers to the mean of the probability over agents. Our $\alpha$ set to $\{ 0.1, 0.01, 0.001, 0.0001, 0.00001\}$ in the population.

\paragraph{RR} RR shares similar ideas with reward randomization~\cite{tang2021discovering}. The intuition behind our reward randomization is that the reward consists of a bonus and a penalty. Typically, the absolute value of the bonus and penalty is equal. However, if we set the bonus higher and the penalty lower, The trained agent will be more risk-seeking and vice versa. Let parameter $\beta$ to control the reward-shaping, the bonus is $\textit{BONUS} * \beta$, the penalty is $\textit{PENALTY} / \beta$. Our $\beta$ set to $\{0.25, 0.5, 1.0, 2.0, 4.0\}$ in the population. 

\paragraph{PSRO} PSRO involves calculating the payoff matrix among all policies in the policy pool, then determining the Nash equilibrium of this matrix for opponent sampling. Next, PPO is used to update the policy, and the updated policy is added to the policy pool. We use fictitious play with 1000 iterations to approximate the Nash equilibrium.

\section{Additional Results}  \label{app:additional}
\subsection{Win-rates matrix among different risk preferences}
\label{app:matrix}
\begin{figure}[!h]
     \centering
     \subfigure[]{
        \includegraphics[width=0.4\textwidth]{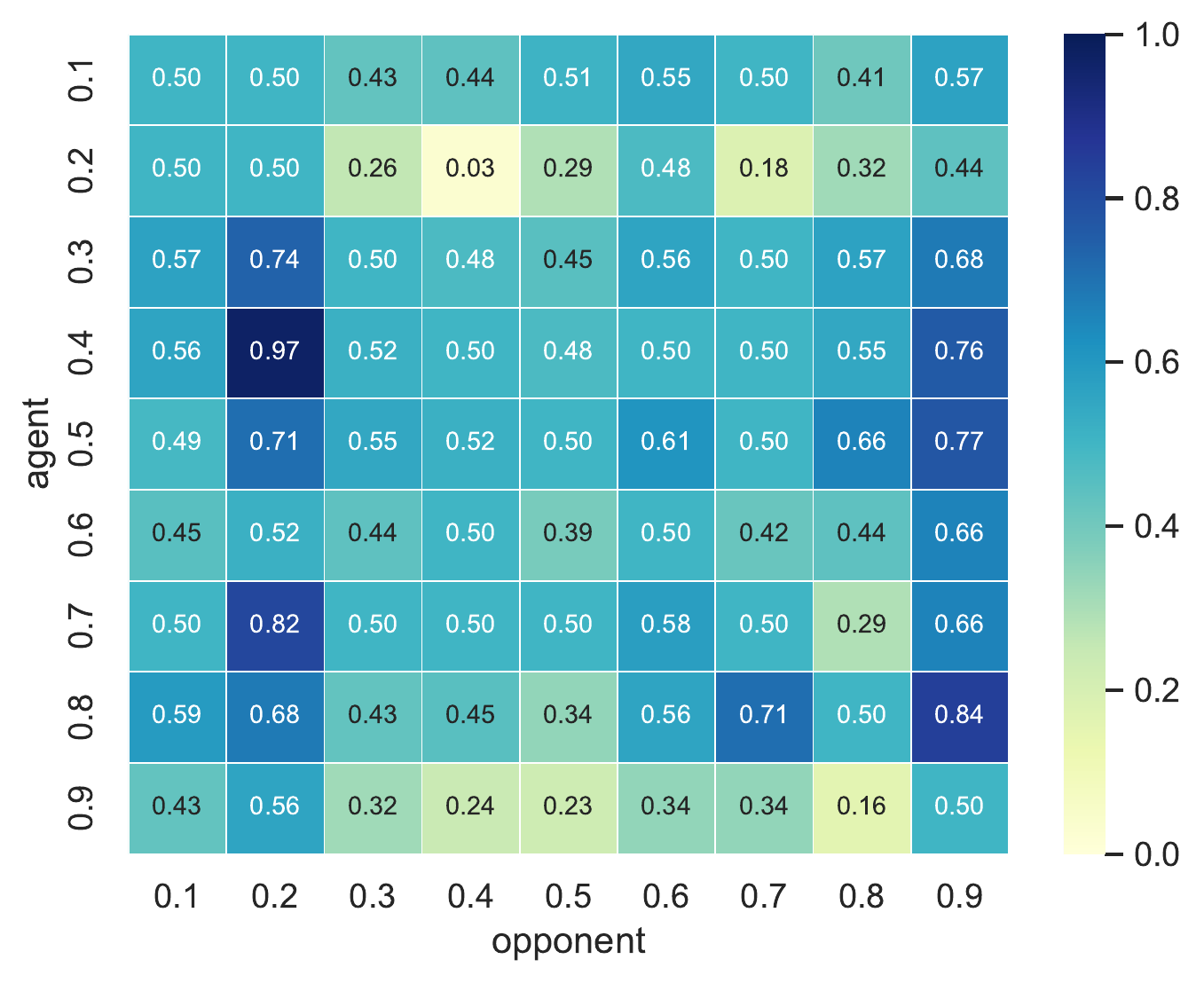}
        \label{fig.volleyball.matrix} 
     } 
     \subfigure[]{
     \includegraphics[width=0.4\textwidth]{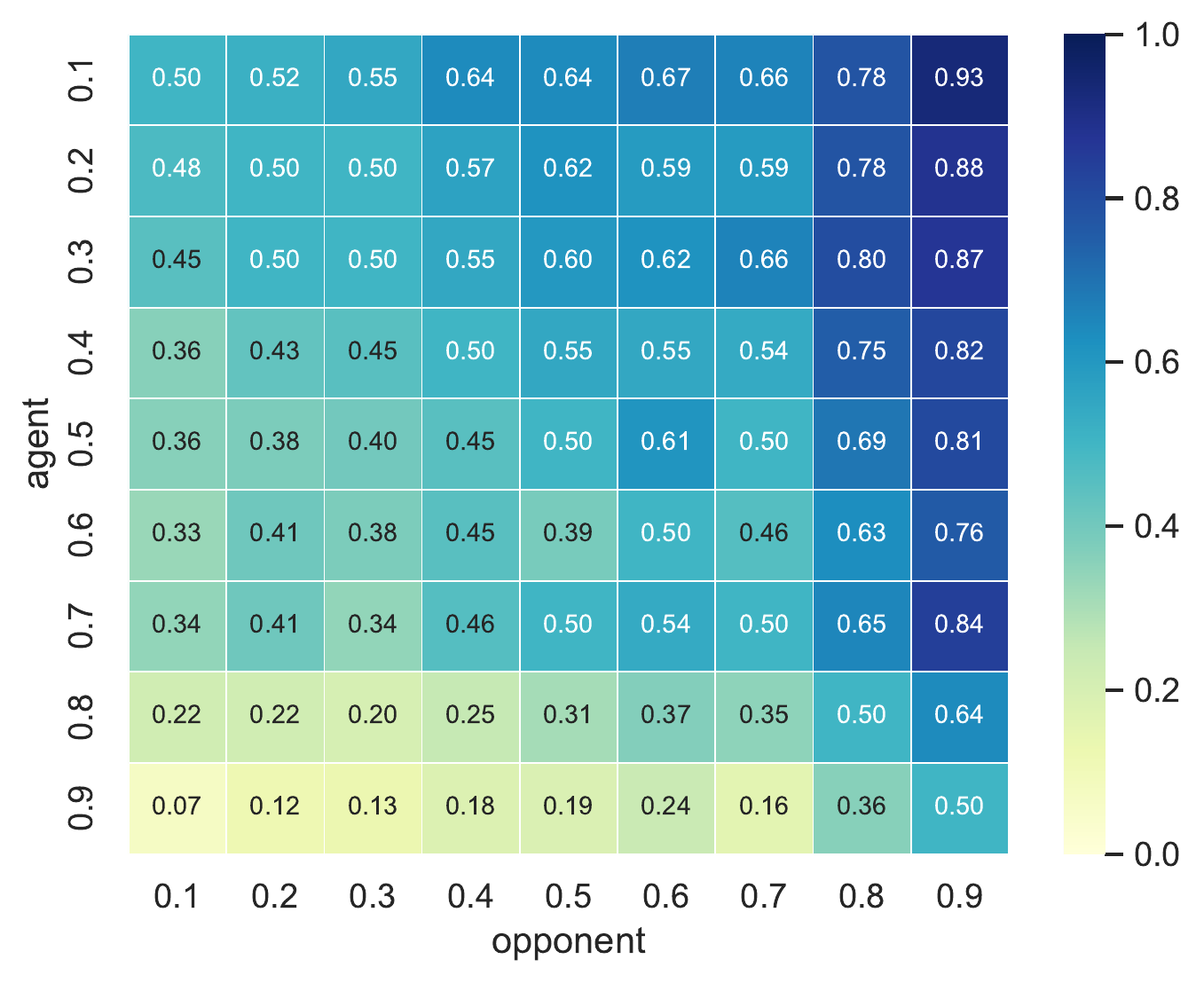}
     \label{fig.ant.matrix}
     }
     \caption{(a) Win-rates matrix among RPPO agents with different risk levels from row players perspective in Slimevolley. A cycle exists among $\tau=0.2, 0.4, 0.6$,  where $\tau=0.4$ completely outperforms $\tau=0.2$, but $\tau=0.2$ can make a tie with $\tau=0.6$ and $\tau=0.6$ can make a tie with $\tau=0.4$. (b) Win-rates matrix among RPPO agents with different risk levels from row players perspective in Sumoants. Risk-averse agents outperform risk-seeking agents.}
     \label{fig:matrix}
\end{figure} 
We trained 9 RPPO agents via self-play, using $\tau \in \{0.1, 0.2, ...0.9\}$ respectively, and the opponents were uniformly sampled from the agent's historical versions. The larger $\tau$ means the more risk-seeking the agent. We pit the 9 agents against each other and compute the expected win-rate over 200 rollouts for each play. The results for both Slimevolley and SumoAnts are shown in the Fig.\ref{fig:matrix}.

The results for Slimevolley reveal that there is not a consistent winner. Some cycles exist, for example, the win-lose relationship among $\tau=0.2, 0.4, 0.6$. Specifically, Agent $\tau=0.4$ against agent $\tau=0.2$ achieved an amazing win-rate of 0.97, which has not occurred in any other plays of $\tau=0.4$. That means $\tau$ = 0.2 get stuck into a local optimum, where $\tau=0.4$ finds its weakness and completely restrains it. However, $\tau=0.2$ is not weak, it can make a tie with $\tau=0.6$, And $\tau=0.6$ is not inferior to the $\tau=0.4$. The existing circles demonstrate the ability of our RPPO method to learn diverse policies in self-play.

As for SumoAnts, we observe a distinct separation if we choose the diagonal as the boundary. It means $\tau<0.5$ outperforms the policy generated by $\tau>0.5$. In addition, agent performance grows monotonically as $\tau$ decreases, implying that more risk-averse agent will perform better in this game. The diverse performance of different risk levels reflects the ability of RPPO to learn diversity.

\subsection{Ablations} \label{app:abla}
 \begin{table*}[!ht]
    \centering
    \subtable[Slimevolley]{
    \resizebox{0.9\linewidth}{!}{
    \begin{tabular}{lccccccccccc}
        \toprule
        \diagbox{Play A}{Play B}      & RPBT    & $\text{RPBT}_{static} $  & $\text{RPPO}_{0.1}$ & $\text{RPPO}_{0.2}$  & $\text{RPPO}_{0.3}$ & $\text{RPPO}_{0.4}$ & $\text{RPPO}_{0.5}$ & $\text{RPPO}_{0.6}$ & $\text{RPPO}_{0.7}$ & $\text{RPPO}_{0.8}$ & $\text{RPPO}_{0.9}$ \\
        \midrule
        RPBT       & -      & 53\%            & 53\%    & 63\%   & 60\% & 68\% & 59\% & 89\% & 70\% & \textbf{96\%} & 96\% \\
        $\text{RPBT}_{static} $  & 47\%     & -   & 78\%    & 92\%   & 75\% & 67\% & 78\% & 91\% & 67\% & \textbf{36\%} & 99\% \\
        \bottomrule
    \end{tabular}}
    }
    \subtable[Sumoants]{
    \resizebox{0.9\linewidth}{!}{
    \begin{tabular}{lccccccccccc}
        \toprule
        \diagbox{Player A}{Player B}     & RPBT      & $\text{RPBT}_{static} $  & $\text{RPPO}_{0.1}$ & $\text{RPPO}_{0.2}$  & $\text{RPPO}_{0.3}$ & $\text{RPPO}_{0.4}$ & $\text{RPPO}_{0.5}$ & $\text{RPPO}_{0.6}$ & $\text{RPPO}_{0.7}$ & $\text{RPPO}_{0.8}$ & $\text{RPPO}_{0.9}$ \\
        \midrule
        RPBT     & - & 49\%      & 50\%    & 51\%   & 56\% & 58\% & 73\% & 75\% & 82\% & 72\% & 97\% \\
        $\text{RPBT}_{static}$ & 51\%  & - & 49\%    & 50\% & 58\%   & 58\% & 69\% & 79\% & 72\% & 78\% & 96\% \\
        \bottomrule
    \end{tabular}}
    }
    \caption{Ablation study on risk levels. RPBT with auto-tuning risk levels performs more robust.}
    \label{tab:risklevel}
\end{table*}

\begin{figure*}[!ht]
    \centering
    \subfigure[Slimevolley]{
    \includegraphics[width=0.45\textwidth]{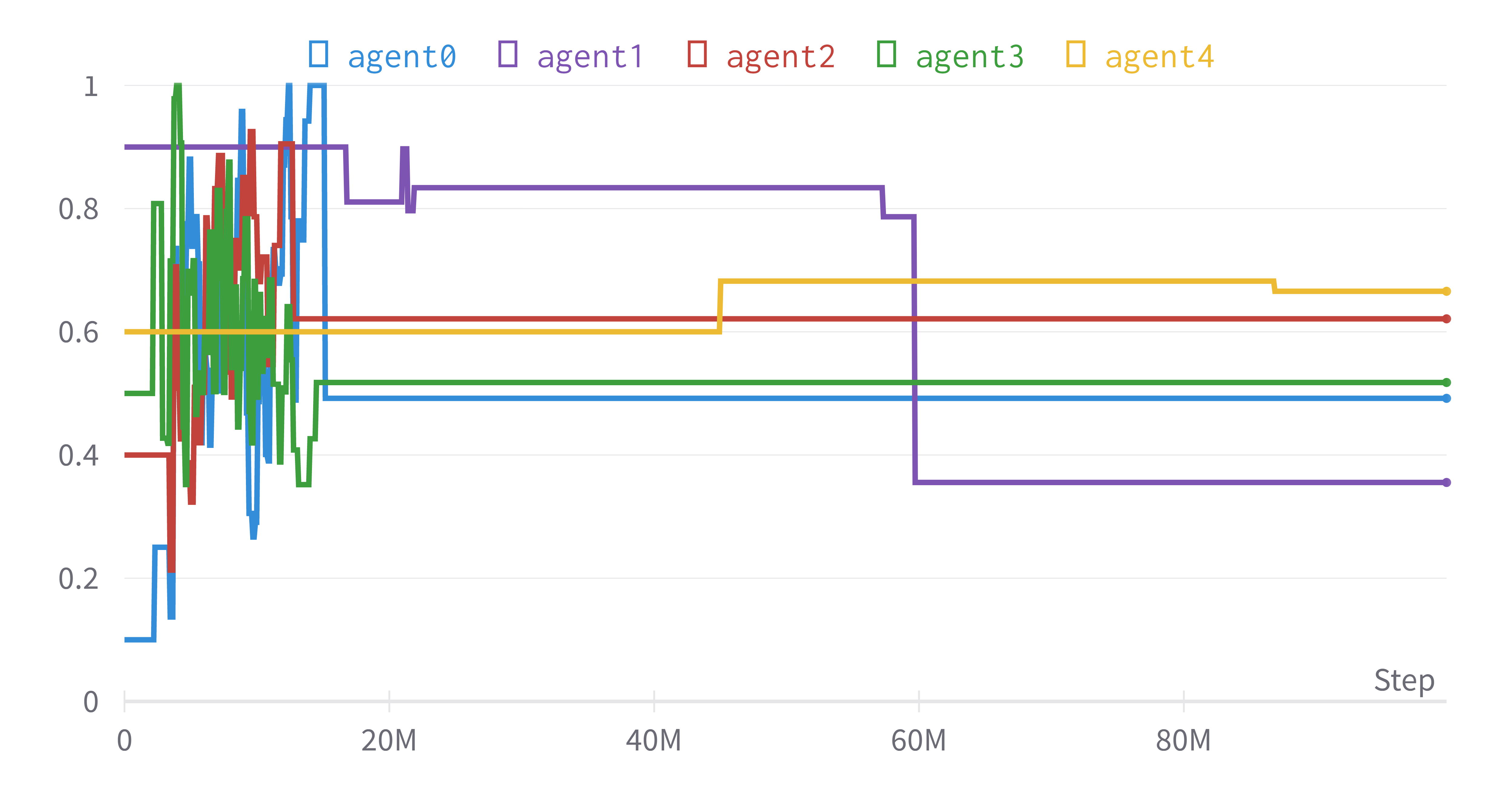}
    }
    \subfigure[Sumoants]{
    \includegraphics[width=0.45\textwidth]{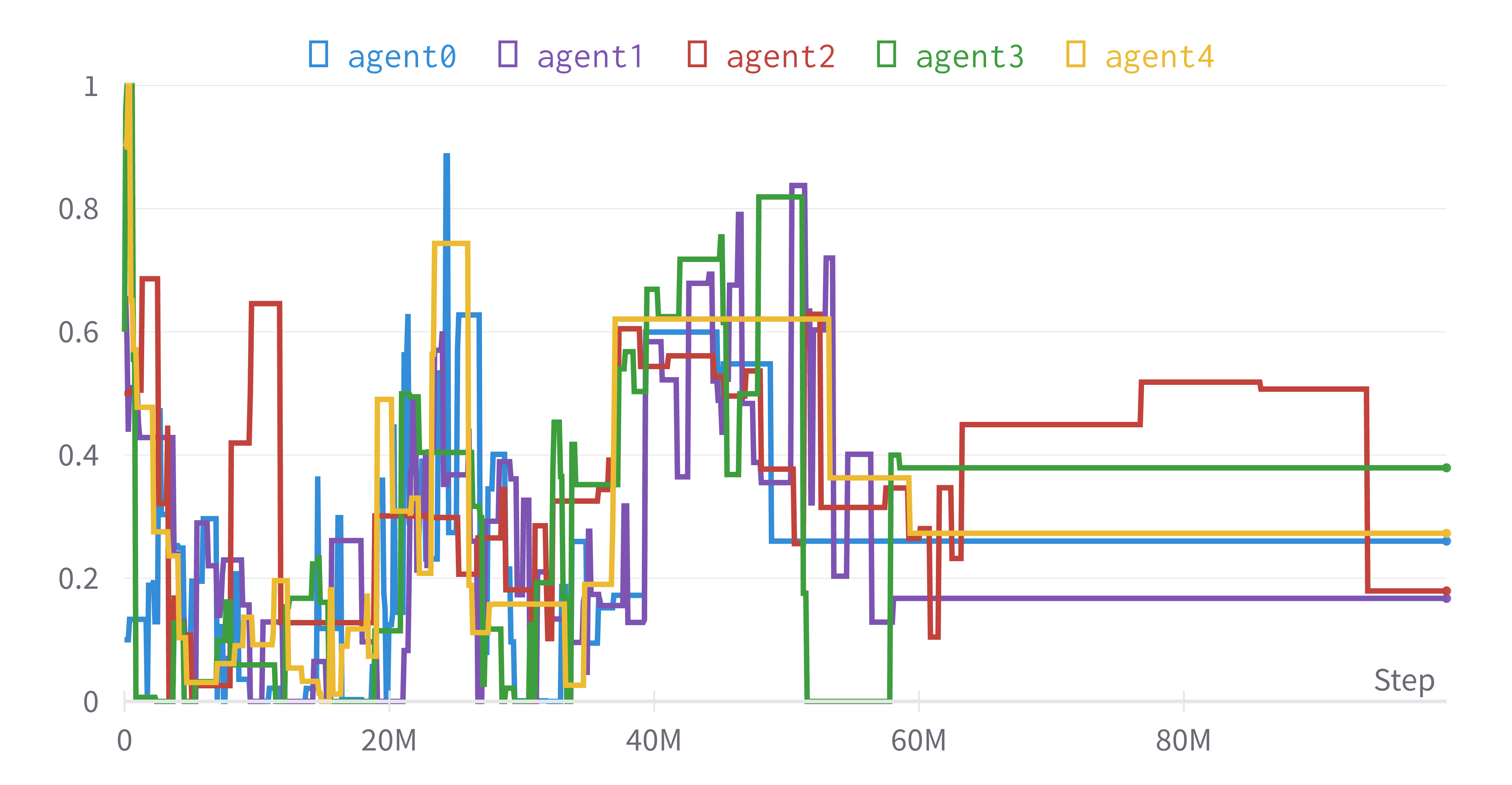}
    }
    \caption{The dynamics of risk level $\tau$ throughout the RPBT training.}
    \label{fig:tau_dynamic}
\end{figure*}

To further study how the auto-tuning risk level takes effect, we compare RPBT with $\text{RPBT}{static}$, which freezes the risk level of each agent throughout the training, and with $\text{RPPO}{\tau}$, which uses a fixed risk level ranging from 0.1 to 0.9. Tab.~\ref{tab:risklevel} shows the results of matching different agents playing against each other. In Slimevolley, we observe that RPBT outperforms all opponents, while $\text{RPBT}_{static}$ is defeated by $\text{RPPO}_{0.8}$. This indicates that auto-tuning risk levels can promote the robustness of agents, while static risk levels might lead to suboptimal performance. In Sumoants, RPBT performs comparably to $\text{RPBT}_{static}$. This is due to the fact that Sumoants has fewer strategy cycles, auto-tuning risk levels can only brings little improvement.

Moreover, we illustrate how risk levels adapt throughout the RPBT training process, shown in Fig.\ref{fig:tau_dynamic}. We observe that RPBT explores a wide range of risk levels in the early stage of training, allowing for diverse strategies to emerge. In the late stage of training, RPBT converges to a stable interval of risk levels (0.3, 0.7) for Slimevolley and (0.2, 0,4) for Sumoants), maintaining diversity while avoiding extreme values that might harm performance. This indicates that RPBT can effectively balance the exploration and exploitation of risk levels in the training.

\end{document}